\documentclass[sn-mathphys]{sn-jnl}

\usepackage[T1]{fontenc}
\usepackage[utf8]{inputenc}
\usepackage{subcaption}

\algrenewcommand{\Return}{\State\algorithmicreturn~}

\jyear{2021}%

\theoremstyle{thmstyleone}%

\theoremstyle{thmstyletwo}%

\theoremstyle{thmstylethree}%
\newtheorem{definition}{Definition}%

\title[Interpretability in Symbolic Regression]{Interpretability in Symbolic Regression: a benchmark of Explanatory Methods using the Feynman data set}

\author*[1]{\fnm{Guilherme} \spfx{Seidyo} \sur{Imai Aldeia}} \email{guilherme.aldeia@ufabc.edu.br}

\author[1]{\fnm{Fabrício} \spfx{Olivetti} \sur{de França}} \email{folivetti@ufabc.edu.br}

\affil*[1]{\orgdiv{Center for Mathematics, Computation and Cognition}, \orgname{Federal University of ABC}, \orgaddress{\street{Av. dos Estados}, \city{Santo Andre}, \postcode{09210580}, \state{SP}, \country{Brazil}}}

\abstract{
    In some situations, the interpretability of the machine learning models plays a role as important as the model accuracy. Interpretability comes from the need to trust the prediction model, verify some of its properties, or even enforce them to improve fairness. Many model-agnostic explanatory methods exists to provide explanations for \textit{black-box} models. In the regression task, the practitioner can use \textit{white-boxes} or \textit{gray-boxes} models to achieve more interpretable results, which is the case of symbolic regression. When using an explanatory method, and since interpretability lacks a rigorous definition, there is a need to evaluate and compare the quality and different explainers. This paper proposes a benchmark scheme to evaluate explanatory methods to explain regression models, mainly symbolic regression models. Experiments were performed using 100 physics equations with different interpretable and non-interpretable regression methods and popular explanation methods, evaluating the performance of the explainers performance with several explanation measures. In addition, we further analyzed four benchmarks from the GP community. The results have shown that Symbolic Regression models can be an interesting alternative to white-box and black-box models that is capable of returning accurate models with appropriate explanations. Regarding the explainers, we observed that Partial Effects and SHAP were the most robust explanation models, with Integrated Gradients being unstable only with tree-based models. This benchmark is publicly available for further experiments.
}


\begin{document}

\maketitle

\section{Introduction}\label{sec:introduction}

The large diffusion of Artificial Intelligence and, more specifically, Machine Learning (ML), has become a reality over the last decades. ML has been seen as a promising application across multiple fields, such as judicial system~\cite{UsingMachineLearningToPredictDecisionsOfTheCourt}, healthcare~\cite{MachineLearningInHealthcareAReview}, scientific discovery in natural sciences~\cite{ExplainableMachineLearningForScientificInsights}, and credit lending~\cite{TowardsExplainableDeepLearningForCreditLending}. 
Many state-of-the-art methods can achieve better performance than experts of the field with high dimensional and non-linear models~\cite{HumanCompetitivenessofGP,Lones2017,8607896,SearchForaGrandTourOfJupiterGalileanMoons,semet2019expert}. Although ML can present high accuracy in several applications, it is arguable that many ML methods generate models that are incomprehensible for a human practitioner. These difficult-to-understand models, also called black-box models, may have undesirable consequences when applied in real-world applications.

The main motivations for explaining predictions and understanding the models are the trust (or lack of) that the model predictions will not lead to catastrophic results; to ensure fairness in the decision process, and to better understand the phenomena under study. The field that studies interpretability is known as \emph{eXplainable Artificial Intelligence} (XAI), where interpretability plays a central role. The recent focus of XAI works was on either improving the accuracy of innate interpretable models or extracting explanations from black-box models.

The motivation for using black-box models is that their accuracy is usually higher than simpler and interpretable models~\cite{DoWeNeedHundredsOfClassifiers}. One way to extract explanations from these kinds of models is to use an explanatory method~\cite{8920138,AUnifiedApproachToInterpretingModelPredictions,PeekingInsideTheBlackBox}. These explanatory methods highlight certain aspects of the model with strategies such as a visual guide of the decision process, human-readable explanation, counterfactual explanation, perceived importance of each feature, among others. Particularly regarding feature importance, an explanatory method can return the aggregated importance of each feature to the decision process, the importance of the features for a single prediction, and the behavior of the feature importance as a function of its own value.

Alternatively to using black-box models and explaining them with explanatory methods, we can use more interpretable models that still provide accurate predictions. One such example is the Symbolic Regression (SR)~\cite{MeasuringFeatureImportanceOfSymbolicRegressionModels}, a regression method that searches for analytical models that best fit the training data. SR is a well-established field that presents competitive performance when compared to state-of-the-art regression algorithms, especially in some domains where a mathematical expression can describe the studied phenomena (\textit{i.e.}, natural science domain). The downside is that sometimes SR is performed with a higher computational cost than traditional regression techniques~\cite{WhereAreWeNow,ContemporarySymbolicRegressionMethods}.


Therefore, SR can be further explored in the XAI field when considering the regression task. Previous works studied the application of SR in XAI tasks, such as~\cite{MeasuringFeatureImportanceOfSymbolicRegressionModels}, where the authors introduce a model-specific explainer for Symbolic Regression models, exploiting the fact that it returns an analytical model. Another benefit of working with analytical models is the possibility of enforcing certain properties to the generated model, such as monotonicity, convexity, or symmetry~\cite{ShapeConstrainedSymbolicRegressionImprovingExtrapolationPriorKnowledge}. Different methods have been proposed to improve performance and interpretability of the SR results~\cite{GainingDeeperInsightsInSymbolicRegression}, and SR has a potential of obtaining easier-to-interpret equations~\cite{SymTree, GainingDeeperInsightsInSymbolicRegression, ApplyingGeneticProgrammingToImproveInterpretability,InteractionTransformationEvolutionaryAlgorithmSymbolicRegression,InteractionTransformationSymbolicRegressionExtremeLearningMachine,SimulatedAnnealingSymbolicRegression}. In the interpretability context, there are works that uses Symbolic Regression as an explanatory method, being used to generate simpler models to explain black-box predictions~\cite{ApplyingGeneticProgrammingToImproveInterpretability, ExplainingSRPredictions}.


One problem with SR is that, while some authors claim it as being more interpretable than black-box approaches, the interpretability is usually measured as the size of the generated expressions \cite{ASurveyOfMethodsForExplainingBlackBoxModels}, a vague notion of interpretability that only suits methods that return mathematical expressions. Because of that, there is still a gap between the promising application of SR in contexts where interpretability plays a central role. Thus, the explanatory methods could be more explored.

This paper attempts to provide insights and to investigate the quality of interpretability in the context of Symbolic Regression, extending the work in~\cite{MeasuringFeatureImportanceOfSymbolicRegressionModels}. The main objective is to investigate the interpretability of Symbolic Regression models when compared with other regression methods by using different feature importance explanatory methods.

For this purpose, we evaluate two different Symbolic Regression algorithms and compare them to other regression methods in the interpretability spectrum --- from a white-box linear regression to a black-box Multi-Layer Perceptron Neural Network --- with multiple explanatory methods. We created synthetic data sets from known physical equations as a proxy to train and evaluate the ML and explanatory methods. Several measures in the literature are revisited and adapted to measure interpretability quality in the experiments. As a by-product, we make all the source code available in the form of a python module called \textit{iirsBenchmark}\footnote{Open source module available at \url{https://github.com/gAldeia/iirsBenchmark}.}.

The \textit{iirsBenchmark} creates a unified experimental design to generate results to compare interpretability measures with different regression methods. We expect to achieve the following objectives:

\begin{description}
    \item[\textbf{O1}] Provide an evaluation method to assess the performance of different explanatory methods for the regression task through defining and evaluating different quality measures;
    \item[\textbf{O2}] Compare two different symbolic regression methods against many popular regression methods that range from the white-box to the black-box spectrum of interpretability.
\end{description}

In summary, our contributions to the scientific community are \textit{i)} the \textit{iirsBenchmark}, an open-source framework to measure explanations quality that unifies the interface of many popular explanatory methods; \textit{ii)} the revision and proposal of measures to quantitatively evaluate the quality and robustness of explanations; \textit{iii)} a robust background to lay the groundwork of interpretability in the symbolic regression context; and \textit{iv)} an extensive comparative study of explanatory methods, using several regression methods and two hand-picked symbolic regression methods. We expect our work to highlight the benefits of using symbolic regression as an alternative to black-box methods, providing a framework to evaluate explanatory methods with special attention in the symbolic regression context. We discuss the evaluation of explanatory methods and propose a framework expected to help researchers present new and more robust methods by providing new perspectives when evaluating, reporting, and analyzing the results.

The remainder of this paper is organized as follows. Section~\ref{sec:relatedWork} presents a revision focusing on benchmarking symbolic regression and explanatory methods. Section~\ref{sec:background} presents the theoretical background to the interpretability field in symbolic regression. Section~\ref{sec:explanatoryMethods} presents popular feature importance explanatory methods for the regression task. Section~\ref{sec:measuringexplanationsquality} presents measures to evaluate the quality and robustness of explanations. Section~\ref{sec:methods} presents the experimental methods and the python package used in this paper. Section~\ref{sec:results} reports the results, which are discussed in Section~\ref{sec:discussion}. Finally, Section~\ref{sec:conclusion} revisits the objectives and concludes the work, summarizing the findings and pointing out new directions.

\section{Related work}~\label{sec:relatedWork}

The term \textit{black-box} model is widely adopted to describe ML models that are complex and lack transparency. There are different definitions for what a black-box model is: a component that does not reveal anything from its inner design, structure or implementation \cite{PeekingInsideTheBlackBox} (or it reveals the structure, but is too complex to understand~\cite{ASurveyOfMethodsForExplainingBlackBoxModels}); a model derived purely from data with no knowledge about its inner working \cite{PerspectivesOnSystemIdentification}; a model that is not inherently interpretable (\textit{i.e.} they need \textit{post-hoc} explanatory methods); or a model that is incomprehensible to humans or proprietary \cite{InterpretabilityAndExplainabilityAMLZoo, StopExplainingBlackBoxMachineLearningModels}. All different definitions relate to how well it is possible to understand the resulting model without additional tools (is it "transparent" enough to let us see inside? Do its individual components make sense? Can a human simulate the decision process of the model, or be able to predict its output by looking at it?). In contrast, \textit{white-box} models are defined as a: model that can be decomposed into individual parts with an explicit meaning on the problem domain~\cite{OnMembershipofBlackorWhiteBoxofANN, BlackBoxvsWhiteBoxPraticalPointofView}; or a model that does not need any external processes to determine the meaning of its decision process~\cite{ExplainingSRPredictions, InterpretabilityAndExplainabilityAMLZoo}.

However, why do we need to understand high-performance ML models? Understanding its behavior can help detect wrong assumptions derived from biased data, understand the decision process, and gain insights into the domain problem. For example, in~\cite{machineBias} the authors discuss a judicial system that associated a higher likelihood of black people committing future crimes; and in~\cite{AutomatedExperimentsonAdPrivacySettings} the authors proposed automated experiments to detect biased models, which helped them find a biased behavior in a job recommender system, that suggested lower-paying roles to women. Many other examples were reported in~\cite{ASurveyOfMethodsForExplainingBlackBoxModels} showing unfair and discriminatory models against minorities and women.

A debate has started about the relevance of understanding the underlying process guiding the decisions of black-box models, both in the scientific community as well as in the public setting~\cite{PeekingInsideTheBlackBox}. 

Nevertheless, the term \textit{interpretability} still lacks a formal definition, even though the intuitive definition can be reasonable. In~\cite{TheMythosOfModelInterpretability} the authors noticed that in many works, the terms are used without specifying their meaning. Some authors adopt \textit{interpretability} and \textit{explainability} interchangeably~\cite{MachineLearningInterpretabilityASurveyOnMethods}. Interpretability is seen as: a passive characteristic of a model, in the sense of how much it makes sense to a human~\cite{XAIConceptsTaxonomiesOpportunitesAndChallenges}; the ability to explain or present in comprehensible ways the model to a human~\cite{TowardsARigorousScienceOfInterpretableML, InterpretabilityAndExplainabilityAMLZoo}; a domain-specific notion, where an interpretable model is useful to someone or obeys domain knowledge~\cite{StopExplainingBlackBoxMachineLearningModels}; the description of the inner functioning of the system in a comprehensible way to humans~\cite{ExplainingExplanations}; or the ability to explain or provide meaning without the necessity of additional information~\cite{ASurveyOfMethodsForExplainingBlackBoxModels}.

To improve interpretability, one can opt to use white-box models, those considered intrinsically interpretable. Examples are linear models, decision trees, and decision rules~\cite{StopExplainingBlackBoxMachineLearningModels}. The downside of using simple and interpretable methods is that they sometimes present a less accurate model when compared to black-box models~\cite{DoWeNeedHundredsOfClassifiers}. Between the white-box and black-box endpoints of the interpretability spectrum, several models cannot be classified as either of those, also called gray-box models, sometimes seen as a compromise between high accuracy and more interpretable models --- although some authors, for example, in \cite{StopExplainingBlackBoxMachineLearningModels}, argue that there is not necessarily a trade-off between them. There is also a belief that complex models are more accurate, which does not hold in every situation. The use of more interpretable models should be preferred when interpretability is a concern. Notice that a white/gray/black-box classification is sometimes subjective, and any white-box model can become gray or black-box depending on the situation, for example, a linear model with many features or a decision tree for a regression problem.

Another possibility is to use a \textit{post-hoc Explanatory Method}, that returns an \textit{explainer model} used to generate \textit{explanations} to help understanding an ML model. The explanation can be a feature importance attribution, visual, prototype, or counterfactual explanation. The explanation is used to work around the lack of transparency of the model. One popular explanation is the feature importance which usually returns a vector with numerical values indicating how much each feature contributes to the model prediction. Despite its popularity, it is still an open question how to validate the performance of feature importance explanatory methods since interpretability lacks a rigorous definition \cite{TheMythosOfModelInterpretability}.

While interpretability is an important concern and an active field of study, we also notice that some authors argue otherwise. In \cite{TheHumanBodyisaBlackBox}, the authors questions if interpretability in the healthcare context is necessary, then designed, proposed, and evaluated an ML tool that had other mechanisms besides interpretability to provide a trustable and accountable system. In \cite{TheFalseHopeofCurrentApproaches} the authors criticize the quality of the current explanatory methods. They argue that evidence already shows existing biases toward over-trusting computer systems --- a system that provides explanations can increase the confidence in it, thus leading to decreased vigilance and auditing of these systems.

Even though there are many explanatory methods proposed recently, some works in the literature report results showing that explanatory approaches do not provide a better understanding of the problem. In \cite{ReadingRace}, the authors try to explain deep learning models using explanatory methods, which point to all regions of the image and relies on non-trivial proxies, being unable to explain the model behavior. To obtain more robust explanatory methods, we first need to provide a proper evaluation methodology, which is considered an open question in the field.
Although many feature importance methods were proposed and widely adopted, literature reports several problems on using some of the state-of-the-art interpretability methods, as stated in many recent works~\cite{StopExplainingBlackBoxMachineLearningModels, BenchmarkingAttributionMethodsWithRelativeFeatureImportance, TheStrugglesofFeatureBasedExplanations, TheDangersOfPostHocInterpretability, CanITrustTheExplainer, OnTheRobustnessOfInterpretabilityMethods, PleaseStopPermutingFeatures, OnEvaluatingCorrectnessofXAI}. To cite some, in~\cite{CanITrustTheExplainer}, experiments have shown that explainers are prone to select non-important features as the most important feature. In~\cite{OnTheRobustnessOfInterpretabilityMethods} the authors compared different explanatory methods, finding that state-of-the-art explainers are sensitive to irrelevant variations in the input data, resulting in very different explanations. Authors in~\cite{PleaseStopPermutingFeatures} made criticisms to a class of explanatory methods based on permutations, showing that complex ML methods learn correlation structures that are disrupted when permuting values. Permutation-based methods break the relationship of the features on the black-box model, presenting wrong explanations. In~\cite{OnEvaluatingCorrectnessofXAI} the authors have shown that different feature importance explanatory methods return different and opposing explanations for similar inputs.

Wrong explanations can have multiple causes, such as a misprediction of the black-box model (we must assume that the prediction model is not perfect), an inaccurate explainer (due to lack of data around a specific point), among others. This means that evaluation of explanations must consider that poor performance is not necessarily related to the explainer but also the explained model. In~\cite{TheDangersOfPostHocInterpretability}, the authors argue that the explanatory methods can rely on learned artifacts from the black-box model instead of actual knowledge learned from the data. Even if the explanatory method is able to generate flawless explanations, we notice that the prediction model can use non-rational correlations. For example, in~\cite{IntelligibleModelsForHealthCare} authors inspected an interpretable rule-based algorithm for predicting pneumonia death risk and found that their model presented a high correlation between patients with asthma having a more negligible death risk, and, after some investigation, they found a causal relationship between the medicine used by asthma patients and the lower risk of death. Authors in~\cite{OnEvaluatingCorrectnessofXAI} show that classification accuracy positively correlates with explanation accuracy. Many other works~\cite{OnEvaluatingCorrectnessofXAI, GeneralPitfallsOfModelAgnosticInterpretationMethods, PleaseStopPermutingFeatures} have shown that learned interactions of the black-box model could not be correctly handled by the explainers, leading to wrong explanations.

It is still considered an open question how to validate the faithfulness of the explanations concerning the explained model~\cite{CanITrustTheExplainer}. In addition, the lack of data with available ground-truth makes it hard to quantitatively measure explanation quality~\cite{TowardsQuantitativeEvaluationOfInterpretabilityMethods, TheStrugglesofFeatureBasedExplanations, MeasuringFeatureImportanceOfSymbolicRegressionModels}, making experiments prone to experimenter biases. In order to evaluate the correctness of explanations, having a ground-truth is essential. Several works in the literature explores and proposes different ways of generating synthetic data sets to evaluate explanations~\cite{OnEvaluatingCorrectnessofXAI,  BenchmarkingAttributionMethodsWithRelativeFeatureImportance, PleaseStopPermutingFeatures, EvaluatingLocalExplanationMethodsOnGroundTruth, GeneralPitfallsOfModelAgnosticInterpretationMethods, MeasuringFeatureImportanceOfSymbolicRegressionModels}. When working with real-world data~\cite{ABenchmarkForInterpretabilityMethodsInDeepNeuralNetworks, TowardsQuantitativeEvaluationOfInterpretabilityMethods, CanITrustTheExplainer, TheDangersOfPostHocInterpretability, OnTheRobustnessOfInterpretabilityMethods, TheStrugglesofFeatureBasedExplanations}, the ground-truth is established heuristically. Another important view of this issue is that, since explanations are meant to be used by humans, there should be a human-centered evaluation for such explanations. Some research groups argue that the XAI field should actively work together with Human-Computer Interface and Social Science fields to move towards intelligible interpretations of prediction models~\cite{vaughan2020human}.

Regarding symbolic regression methods, prior works~\cite{BetterGPBenchmarks, GeneticProgrammingNeedsBetterBenchmarks} raised different questions concerning the evaluation of symbolic regression in the literature, pointing out problems such as the usage of toy data sets; lack of reproducibility; and poor experimentation, statistics, and reporting. In \cite{BetterGPBenchmarks} the authors surveyed the Genetic Programming (GP) community and presented some points to consider when designing symbolic regression benchmarks. Apart from that, one of the promising applications of symbolic regression is discovering new physics equations and describing systems by finding appropriate expressions. Introduced recently in \cite{AIFeynman} and further investigated in \cite{AIFeynman2dot0}, the Feynman benchmark comprehends 100 problems that can be explored by symbolic regression, which has been proven to be a complex task to solve these problems by finding the exact expression without domain-specific improvements to current algorithms, providing a challenging task such as the other GP problems, but with real-world physics equations.

\section{Symbolic Regression}~\label{sec:background}

\textit{Regression Analysis} is the task of estimating the expected value of a dependent feature $Y$, also known as target-value or outcome, conditioned to a set of independent features~\cite{StatisticalLearningFromARegressionPerspective}. The main goal is to understand the structural relationships between the dependent and the independent features.
It is based on the assumption that an unknown function $f(\mathbf{X})=Y$ describes the relationship between the independent and dependent features. From a modeling perspective, many regression algorithms parts from some assumptions about this relationship and chooses an appropriate model $\widehat{f}(\mathbf{X}, \beta)$, where $\beta$ is adjusted to minimize the approximation error of the model\cite{RegressionModelingStrategies}.

A simple and well-known model is the linear regression, that predicts an outcome with the linear function:

\begin{equation}
    \widehat{f}(\mathbf{X},\beta) = \beta_0 + \mathbf{X}^T\boldsymbol{\beta}.
\end{equation}

The interpretation of this model can be straightforward in the context of feature importance: we can say that by increasing $X_i$ in one unit, the prediction will increase in $\beta_i$ units; or that the $i$-th feature contributed with $\beta_i X_i$ to the outcome from the reference prediction $\beta_0$.

Despite the straightforward interpretation, the linearity assumption limits the achievable accuracy of this model for many data sets. This can be alleviated with manual feature engineering by adding feature interaction and non-linearity. Another possibility is to use one of the linear model generalizations such as Generalized Linear Models and Generalized Additive Models.

Symbolic Regression (SR) is another alternative to generating a regression model. It usually starts from a randomly initialized free-form mathematical expression and performs the search and optimization of both model structure and its inner coefficients \cite{SymbolicRegressionBasedHybridSemiparametric}.
The search space of SR is constrained by how the mathematical expressions are represented and the primitive set of mathematical structures~\cite{SymTree}. This approach has more potential to find interpretable solutions than black-box approaches, as argued by many authors of the field~\cite{SymTree, ExplainingSRPredictions, GainingDeeperInsightsInSymbolicRegression, ApplyingGeneticProgrammingToImproveInterpretability}.

The most common representation is the expression tree, where each node is either a \textbf{leaf node}, holding a constant value or feature symbol; or an \textbf{inner node}, representing an $n$-ary function $f$, followed by $n$ child nodes. For a given observation $\mathbf{X}$, the terminal symbols are evaluated as the constant value it holds, or as the value that a feature $X_i$ assumes on $\mathbf{X}$. The inner nodes are evaluated as the application of the $f$ function over its $m$ children $t_1, t_2, \ldots, t_m$, returning $f(t_1, t_2, \ldots, t_m)$. 

Symbolic Regression can be seen as an optimization problem in which we want to find the sub-optimal expression $g*$ that minimizes a cost function from the set $\mathbf{G}$ of all representable mathematical expressions.

\subsection{Genetic Programming for Symbolic Regression}~\label{subsubsec:cannonicalSymbolicRegression}

Genetic Programming is an evolutionary algorithm commonly used for Symbolic Regression~\cite{GPOnTheProgrammingOfComputersByMeansOfNaturalSelection}.
The search for the sub-optimal expression starts with a random population of expression trees, created using a primitive set with functions nodes $\mathbf{f}$ and constant nodes $\mathbf{c}$, with tree depth limited by constraints $\Gamma$. While a stop criteria (\textit{e.g.} a fixed number of generations) is not met, the algorithm creates the next generation as children from the current population that will compete to replace the previous individuals with probabilities taken from their fitness evaluated over the training data. When the stop criteria is met, then it returns the best expression of the last existing generation. The whole process is summarized in Algorithm \ref{alg:canonical}.

\begin{algorithm}[t!]
\caption{Canonical Symbolic Regression}\label{alg:canonical}

    \begin{algorithmic}[1]
        \Require Function set $\mathbf{f}$, constant set $\mathbf{c}$, tree constraints $\Gamma$, stop criteria $\Omega$, population size $\textup{p}$, train data $(\mathcal{X}=\{ (x_1, x_2, \ldots, x_d)_i \}_{i=1}^{n}, \mathbf{Y}=\{ y_i \}_{i=1}^{n})$
        \Ensure Symbolic expression $\widehat{f}$
        
        \State pop $\leftarrow$ $[$ \Call{generate\_random\_tree}{$\mathbf{f}$, $\mathbf{c}$, $\Gamma$} for $\_ \in [1, 2, ..., \textup{p}]$ $]$;
        \While{Criteria $\Omega$ is not meet}
            \State parents $\leftarrow$ $[$ \Call{select\_N\_parents}{pop} for $\_ \in [1, 2, ..., p]$ $]$;
            \State children $\leftarrow$ $[$ \Call{crossover}{p1, p2, $\ldots$} for $(\text{p1}, \text{p2}, \ldots) \in \text{parents}$ $]$;
            \State mutants $\leftarrow$ $[$ \Call{mutate}{c} for $\text{c} \in \text{children}$ $]$;
            \State pop $\leftarrow$ \Call{replace}{pop $++$ mutants, $\mathcal{X}$, $\mathbf{Y}$};
        \EndWhile
        \Return arg max $[$ \Call{fitness}{p, $\mathcal{X}$, $\mathbf{Y}$} for $\text{p} \in \text{pop}$ $]$;
    \end{algorithmic}
\end{algorithm}

In this approach, both the function structure and the free parameters are created and adjusted by the simulated evolutionary process. In the original SR implementation, by Koza~\cite{GPOnTheProgrammingOfComputersByMeansOfNaturalSelection}, the algorithm finds values for the free parameters using two different strategies: the Ephemeral Random Constant (ERC), which creates constants by taking values at random from a pre-defined interval; or by using pre-defined constant values (such as $\pi$, Euler's constant or arbitrary numerical values).

Several free parameter adjustment methods were proposed later, with approaches based on \textit{gradient descent} or \textit{hill-climbing} optimization methods, as identified by the literature review in \cite{ParameterIdentificationForSRUsingNonlinearLS}.

\subsection{State-of-the-Art Symbolic Regression}~\label{subsubsec:stateoftheartSymbolicRegression}

In~\cite{kommenda2020parameter}, the authors proposed an extension to the original GP, an algorithm named GP-NLS. In this extension, the authors added two additional steps right before evaluating the fitness of an expression. This implementation is currently available in the Operon C++ Framework \cite{OperoncppAnEfficientGPFramework}. 

In~\cite{ContemporarySymbolicRegressionMethods}, the authors compared different regression algorithms with other contemporary symbolic regression methods in a large set of benchmark problems, including the Feynman data sets presented in \cite{AIFeynman}.Overall, the Operon GP-NLS was the best-ranked algorithm, indicating that it provides the best average performance for different problems.

\begin{figure}[t!]
    \centering
    \begin{subfigure}[b]{0.35\textwidth}
        \centering
        \includegraphics{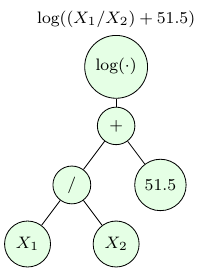}
        \caption{Example of a SR tree.}
        \label{fig:normal_tree}
    \end{subfigure}%
    \begin{subfigure}[b]{0.5\textwidth}
        \centering
        \includegraphics{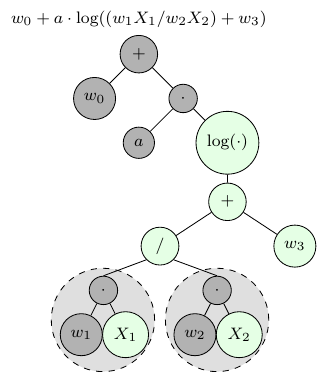}
        \caption{Operon GP-NLS expanded tree.}
        \label{fig:expanded_tree}
    \end{subfigure}%
    \caption{Operon GP-NLS expanded tree (new nodes in gray, original tree nodes in a lighter color). The offset node $w_0$ is summed with the original tree, now scaled with the scale node $a$. Every feature has a coefficient associated with it ($w_1, w_2$). The original coefficient --- a fixed value --- is transformed into a free parameter $w_3$. The Levenberg-Marquardt algorithm will find optimal values for all free parameters in the expression tree ($\{w_0, a, w_1, w_2, w_3\}$)}.
    \label{fig:operonTree}
\end{figure}

The first step expands every individual in the population by adding an offset and a scaling node at the root of the tree, also adding a coefficient to every feature node. The second step adjusts the newly added free parameters by solving a non-linear least-squares problem, using the Levenberg–Marquardt algorithm, adjusting the values of all free parameters. Fig. \ref{fig:operonTree} shows an expression tree before and after the expansion. Notice that this modification does not persist during the crossover and mutation --- the evolutionary operators are applied only to the original tree.

We represent an Operon tree by $M_{g, \theta}$, where $g$ is the differentiable function of the expanded tree $T'$n and $\theta$ is the vector of free parameters to be optimized in $T'$. Let $H: \mathbb{R}^p \rightarrow \mathbb{R}^n$ be a function that evaluates the parameter $\theta$ and returns the difference between the output of the model $M_{g, \theta}$ and the real value for observations $\mathcal{X}$:

\begin{equation}
    H(\theta) = G(M_{g, \theta}, \mathcal{X}) - \mathbf{Y},
\end{equation}

\noindent with $G:\mathbb{R}^{n \times d} \rightarrow \mathbb{R}^n$ being a function that evaluates the prediction of the model $M_{g, \theta}$ for the observations in $\mathcal{X}$.

The optimal coefficients can be found using the Jacobian matrix of $H$, through an interactive process of gradient descent, in which each interaction uses a step $\Delta \theta$ obtained by the linearization $H(\theta + \Delta\theta) \approx H(\theta) + J(\theta)\Delta\theta$. This process can be seen as a non-linear optimization problem.

The Operon GP-NLS algorithm follows the same steps as in Alg.~\ref{alg:canonical}, with the addition of creating and adjusting the free parameters. The initial population is created using the PTC2 initialization method \cite{PTC2}, which allows controlling both depth and number of nodes in the generated trees while favoring balanced trees.

\subsection{Interaction-Transformation representation}~\label{subsubsec:differentRepresentations}

One way to improve the simplicity of the models found by SR algorithms is to constrain the search space to function forms, thus excluding complex structures such as function chaining. The \emph{Interaction-Transformation}(IT) representation~\cite{SymTree} was proposed to constrain the search space to an affine combination of non-linear transformations applied to interaction terms.

For a problem with $d$ input features $\mathbf{X} = (X_{1}, X_{2}, \cdots, X_{d})$ the IT representation is any regression model of the form:

\begin{equation} \label{eq: ExprIT}
    \hat{f}(\mathbf{X}, \beta) = \beta_0 + \sum_{j=1}^{t}{\beta_j \cdot g_j(p_j(\mathbf{X}, \mathbf{k}_j))},
\end{equation}

\noindent where $\beta_j \in \mathbb{R}$ is the coefficient of the $j$-th term, and $\beta_0$ is the expression intercept. The function $g_j: \mathbb{R} \rightarrow \mathbb {R}$ is any unary function, called transformation function, and $p_j$ is the interaction function, defined as:

\begin{equation}
    p_j(\mathbf{X}, \mathbf{k}) = \prod_{i = 1}^{d} X_i^{k_{i}},
\end{equation}

\noindent with $\mathbf{k}=\{k_1, k_2, \ldots, k_d \} \in \mathbb{Z}^{d}$ is the strength of interaction for the feature $X_i$ in the interaction function index $j$.

In~\cite{InteractionTransformationEvolutionaryAlgorithmSymbolicRegression} the authors proposed the Interaction-Transformation Evolutionary Algorithm (ITEA) to search for the IT expression that best fits the data. The ITEA searches for transformation functions $g_j$ and strengths $k_{ij}$ for each $j$. Ordinary least squares can determine the values of $\beta$.
ITEA follows a similar evolutionary process as in Alg.~\ref{alg:canonical} but applies only the mutation operator to create the children expressions. The mutation operators can expand, shrink, or make local adjustments to an IT expression:

\begin{itemize}
    \item \textbf{Expand:} mutations that add a new pair $(g_i, \mathbf{k}_j)$ to the expression. The new IT term can be a random one or can be the positive/negative element-wise combination of two existing terms on the expression;
    \item \textbf{Shrinkage:} performs the removal of a random existing pair $(g_i, \mathbf{k}_j)$ of the expression;
    \item \textbf{Local modification:} randomly changes one value in the  interaction strengths $\mathbf{k}_j$ for a random $j$ without adding or removing any new structures. 
\end{itemize}


\section{Explanatory Methods}~\label{sec:explanatoryMethods}

Explanatory methods can alleviate the lack of transparency of black-box models when an explanation of the decision process is required. They are called \textit{post-hoc explanatory method} since they are created after fitting the ML model. These methods can also be useful for white-box models in situations where the model becomes unintelligible due to an \emph{a priori} feature engineering or due to the high dimensionality of a data set. The explanatory method usually returns an \textit{explainer model} capable of providing explanations on demand for new examples or summarizing the behavior of the ML model.

One popular strategy for explaining a model is the feature importance explanation. It generates an explanatory function that inputs the predictor and the training data. Then, the function provides explanations by creating a vector assigning a numerical value for each feature, indicating their relative importance to the prediction.
This explanation can be local when the explainer is used for single observations, or global when the behavior of the predictor is summarized as the expected importance of each feature.

\begin{definition}[Local and Global feature Importance Explanations]
    Given a regressor $\widehat{f}: \mathbb{R}^d \rightarrow \mathbb{R}$ trained with a data set $(\mathcal{X}=\{ (x_1, x_2, \ldots, x_d)_i \}_{i=1}^{n}, \mathbf{Y}=\{ y_i \}_{i=1}^{n})$, a \textbf{local explainer} is a model $\psi: \mathbb{R}^d \rightarrow \mathbb{R}^d$ that returns a $d$-dimensional vector where the $i$-th position contains a value with the importance of the $i$-th feature to a particular observation $\mathbf{x} \in \mathbb{R}^d$. Similarly, a \textbf{global explainer} is a model $\phi: \mathbb{R}^{n \times d} \rightarrow \mathbb{R}^d$ that, given a data set $\mathcal{X}$ it returns a $d$-dimensional vector with the aggregated importance of each feature.
\end{definition}

\begin{figure}[t!]
    \centering
    \includegraphics[width=\linewidth]{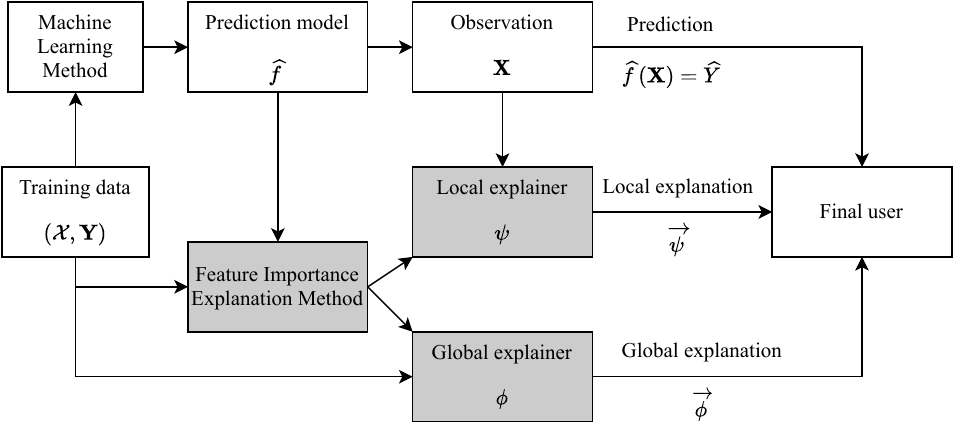}
    
    \caption{Diagram illustrating the relationships between the black-box with its predictions and the explainer with its explanations. The ML prediction model is used as an input together with the training data. Then, it generates feature importance explainers to help understand the model. Adapted from \cite{avaliacaoDaInterpretabilidade}.}
    \label{fig:explanationMethods}
\end{figure}

Fig. \ref{fig:explanationMethods} shows a diagram expressing the relationship between the black-box model, the explanatory method, and their respective outputs. The explanatory model is generated using the training data and the black-box model.

The remaining of this section will present popular explanatory methods for feature importance.

\subsection{Permutation Importance}~\label{subsubsec:permutationImportance}

The Permutation Importance method was first described by Breiman when introducing the Random Forests algorithm \cite{RandomForests}. This method measures how much the coefficient of determination ($R^2$) of the model reduces when we permute the values of a feature on the training data. This permutation is repeated several times to measure the expected variation of $R^2$. The main idea is that the random permutation simulates the removal of this feature and how much that impacts the prediction model.

\begin{definition}[Permutation Importance -- Global Importance Explanation]
    Let $s$ be the $R^2$ score of the prediction model $\widehat{f}$. The importance of a feature $j$ is calculated as:
    
     \begin{equation}
         \phi_j (\widehat{f}, \mathcal{X}) = s - \frac{1}{K} \sum_{k = 1}^{K} \textup{R}^2(\widehat{f}(\tilde{\mathcal{X}}_{k,j}), \mathbf{Y}),
     \end{equation}
    
     \noindent where the $R^2 $ of $\widehat{f}$ is evaluated over $\tilde{\mathcal{X}}_{k, j}$, a copy of the matrix of problem features for all observations $\mathcal{X}$ with the $j$-th column randomly shuffled, and $K$ is the number of iterations where the matrix $\tilde{\mathcal{X}}_{k, j}$ was recreated.
\end{definition}

\subsubsection{Local Interpretable Model-agnostic Explanations (LIME)}~\label{subsubsec:LIME}

LIME \cite{WhyShouldITrustYou} provides a local explanation for a single observation by fitting a linear model using random samples generated within the neighborhood of the observation. The linear model is used to obtain the feature importance. A mathematical formulation of LIME was presented in \cite{AUnifiedApproachToInterpretingModelPredictions}. This method approximates the black-box predictions using an interpretable model and only supports local explanations in the original version.

\begin{definition}[LIME -- Local Importance Explanation]

    Given an observation $\mathbf{x}$ and a predictor $\widehat{f}$, LIME generates a neighborhood with normal distribution $N_{\mathbf{x}}$ and fits a local linear model for a number of $r$ sampled observations $\mathcal{D}' = \{ (\mathbf{x}'_i, \widehat{f}(\mathbf{x}')_i) \}_{i=1}^{r}$. The feature importances are calculated by minimizing the objective function:
    
    \begin{equation}
        \psi(\widehat{f}, \mathbf{x}) = \underset{g \in G}{\text{arg min }} L(\widehat{f}, g, \pi_{N_{\mathbf{X}}}) + \Omega (g),
    \end{equation}
    
    \noindent where $L(\widehat{f}, g, \pi_{N_{\mathbf{x}}})$ is a loss function that evaluates how good $\widehat{f}$ is locally approximated by $g$ in the local neighborhood of $\mathbf{x}$, $\pi_{N_{\mathbf{x}}}$ being a kernel that weights the neighborhood with its distance from the original point, $G$ is the set of interpretable linear models, and $\Omega$ is a penalty function measuring the model complexity.
    The local explanation $\overrightarrow{\psi}$ is the vector of coefficients from the local linear model $\xi$.
\end{definition}

\subsubsection{Explain by Local Approximation (ELA)}~\label{subsubsec:ELA}

ELA \cite{miranda2020explaining} is a local explainer based on LIME and was first introduced using a symbolic regression as the regression model. This method also uses a linear model to obtain local explanations for a regressor but, different from LIME, it does not create new samples with a normal distribution around the given observation. Instead, it uses the $k$-nearest neighbors to create the model. While this has the advantage of using real-world data instead of artificially created points, it assumes that we have representative training data. The authors also propose a global explanatory method in the original paper, but it is a visual explanation; thus, it is not considered in this work.

\begin{definition}[ELA -- Local Importance Explanation]
     Given an observation $\mathbf{x}$ and a predictor $\widehat{f}$, ELA generates a subset $\mathcal{N}_{k}$ of the available data $\mathcal{X}$  by collecting the $k$-nearest neighbors of $\mathbf{x}$ using the euclidean distance to fit a local linear model. If the model has an intrinsic feature selection property, the euclidean distance is calculated only with the features selected by the regression model. As in LIME, the coefficients of this linear model represent the importance of each feature.
    
\end{definition}

\subsubsection{SHapley Additive exPlanations (SHAP)}~\label{subsubsec:SHAP}

SHAP \cite{AUnifiedApproachToInterpretingModelPredictions} is an algorithm that estimates the \textit{Shapley} values as a feature importance measure, based on coalition game theory, where players can contribute positively or negatively to a result of a game. To find the total contribution of each player, and since players can interact between themselves, the Shapley values represent the overall contribution that the player presents, based on all possible coalition of teams. The Shapley value indicates how much a feature contributes to the current prediction w.r.t. the average prediction.

\begin{definition}[SHAP -- Local Importance Explanation]
    Let $F$ be the set of all features, and $S \subseteq F$ a subset of the features, and $\mathbf{x}$ an observation of interest. For each feature $j$, the difference in prediction with and without this feature is evaluated.
    This is computed for all possible subsets $S \subseteq F \setminus \{j\}$. Finally, the weighted mean between all possible coalitions is computed, returning the Shapley Value interpreted as the feature importance:

    \begin{equation}
        \psi_j(\widehat{f}, \mathbf{x}) = \sum_{S \subseteq F \setminus \{ j \}} \frac{\lvert S \rvert !(\lvert F \rvert - \lvert S \rvert - 1)!}{\lvert F \rvert !}[ \widehat{f}_{S\cup \{ j \}}(\mathbf{x}_{S \cup \{ j\}}) - \widehat{f}_S(\mathbf{x}_{S}) ].
    \end{equation}
\end{definition}

In the original SHAP framework implementation, the authors calculate the global explanation as the mean of absolute local explanations for all observations in the training data.

\begin{definition}[SHAP -- Global Importance Explanation]
    Let $\mathcal{X}$ be the training data of the predictor. The global importance for a feature $j$ is the mean of absolute local importance for all observations in $D$:
    
    \begin{equation}
        \phi_{j}(\widehat{f}, \mathcal{X}) =  \frac{1}{n} \sum_{i=1}^{n} \lvert \psi_{j}(\widehat{f}, \mathcal{X}_i) \rvert .
    \end{equation}
\end{definition}

\subsubsection{Shapley Additive Global importancE (SAGE)}~\label{subsubsec:SAGE}

In~\cite{SAGE} the authors proposed the SAGE explainer, an extension to SHAP for a global explanation. 
SAGE values measure the uncertainty reduction when incorporating a certain $X_j$ in different subsets of features $\mathbf{X}_S$ to the model, and are interpreted as follows:

\begin{itemize}
    \item The total sum of the values results in the predictive power of the model;
    \item An feature $j$ will have zero effect if $X_j$ is conditionally independent of $\widehat{f}(\mathbf{X})$ given all possible subsets of features in $\mathbf{X}$;
    \item The SAGE values represent the weighted average of uncertainty reduction for the predictions when a feature $X_j$ is considered in the model.
\end{itemize}

\begin{definition}[SAGE global explanation]
    Let a regressor be $\widehat{f}: \mathbb{R}^d \rightarrow \mathbb{R} $ trained with a data set $\mathcal{D} = \{(\mathbf{X}_i, Y_i) \}_{i=1}^{n}$, and $\mathcal{X}$ the matrix of the $d$ problem features for all the $n$ observations. Also, let $F$ be the set of all features in the problem, and $S \subseteq F$ a subset of the features. The SAGE value is calculated the same way as the \textit{Shapley} values but using the  mutual information $I$ of predictions instead of the prediction model:

    \begin{equation}
        \phi_j(\widehat{f}, \mathcal{X}) = \sum_{S \subseteq F \setminus \{ j \}} \frac{\lvert S \rvert !(\lvert F \rvert - \lvert S \rvert - 1)!}{\lvert F \rvert !}[v_f(S\cup \{j\}, \mathcal{X}) - v_f(S, \mathcal{X})],
    \end{equation}
    
    \noindent where $v_f$ is a function that evaluates the predictive power of the subset of features $S$ by comparing the average prediction (absence of all features) with the prediction considering the subset $S$:
    
    \begin{equation}
        v_f(S, \mathcal{X}) = \underbrace{\mathbb{E} \left [ I(f_{\emptyset}(\mathcal{X}_{*, \emptyset}), \mathbf{Y}) \right ]}_{\textup{Average prediction}} - \underbrace{\mathbb{E} \left [ I(f_{S}(\mathcal{X}_{*, S}), \mathbf{Y} ) \right ]}_{\textup{Prediction considering } S},
    \end{equation}
    
    \noindent where $I$ is the mutual information of a model $f_S$ using the features in $S$ and the target values of the training set.
\end{definition}

Both SHAP and SAGE perform expensive computations to estimate the original SHAP value, which is hard to compute thus requires approximations to alleviate the computational burden since the number of permutations grows exponentially with the dimensionality and are estimated by various methods.

\subsubsection{Morris Sensitivity}~\label{subsubsec:morris}

The Morris Sensitivity~\cite{Morris} calculates the importance of each feature based on their \emph{elementary effect} (EE) -- a measurement of how much the output changes with a slight change in the feature value. The values globally represent how sensitive the method is to variations in each feature.
This explainer is implemented in InterpretML framework~\cite{interpretML}. 
\begin{definition}[Morris Sensitivity -- Global Feature Explanation]
    Let the \textit{Elementary effect} be calculated as:
    
    \begin{equation}
        \textup{EE}_j = \frac{\widehat{f}(X_1, X_2, \ldots, X_{j}+\Delta, \ldots, X_n) - \widehat{f}(X_1, X_2, \ldots, X_n)}{\Delta},
    \end{equation}
    
    \noindent where $\Delta$ is the step size, picked in a way that $X_j + \Delta$ is still within the features domain. 
    
    The Morris Sensitivity global importance performs $k$ random paths through the parameter grid, from different starting points in the data set, modifying the features one at a time, computing the mean EE from all trajectories of the feature of interest:
    
    \begin{equation}
        \phi_{j}(\widehat{f}, \mathcal{X}) = \frac{1}{k_j} \sum_{r=1}^{k_j} \textup{EE}_{k}^{r},
    \end{equation}
    
    \noindent where $k_j$ is the number of steps where  $X_j$ has changed.
\end{definition}

\subsubsection{Integrated Gradients}~\label{subsubsec:IG}

Integrated Gradients~\cite{IG} is a local explainer created to calculate an attribution mask for image predictions made by Deep Learning models. The importance of each feature is calculated using the Aumann-Shapley~\cite{aumann2015values} value that extends the Shapley value for infinitely many players. This method also requires that the user specifies a reference point $\mathbf{x}'$ instead of assuming the average prediction. This has the advantage of generating a better explanation for continuous features and using a realistic reference point. On the other hand, it requires calculating the gradient of the model w.r.t. its features.

\begin{definition}[Integrated Gradients -- Local Feature Explanation]

    The Integrated Gradient of the $j$-th feature for a given observation $\mathbf{x}$ is:

    \begin{equation}
        \psi_j(\widehat{f}, \mathbf{x}) = (x_j - x'_j) \int_{\alpha=0}^{1} \frac{\partial \widehat{f}(\mathbf{x}' + \alpha (\mathbf{x} - \mathbf{x}'))}{\partial X_i} d\alpha 
    \end{equation}
    
    \noindent where $\mathbf{x}'$ is the baseline point, and $\alpha$ represents a continuous path between the point being explained and the reference. 
\end{definition}

A Riemman trapezoidal sum approximates the integral, and the derivatives can be obtained numerically by a finite difference method. The values represent the gradients along the path for the predictor to go from the mean prediction (when $\mathbf{x}' = \overline{\mathbf{X}}$ is the mean vector for each feature) to the final prediction. 

\subsubsection{Partial Effects}\label{subsubsec:partialeffects}

The econometric and social science fields commonly use the concept of partial derivatives to investigate the behavior of the independent features and their interactions. There are many works that proposes its usage as an interpretability framework in the context of regression analysis, such as \cite{ggeffects, MarginalEffectsQuantifyingTheEffectOfChanges, UsingPredictionsAndMarginalEffectsToCompareGroups, AGeneralFrameworkForComparingPredictionsAndME, APrimerOnMarginalEffectsI}, to name a few.

The Partial Effect (PE)\footnote{Also called \textit{Marginal Effects} in the literature, but this term can be misleading, as mentioned in \cite{SimpleWaysToInterpretEffectsInModeling}.} measures how an infinitesimal change (or a discrete change of one unit, for discrete features) in an independent feature affects the dependent feature when its co-features are fixed on specific values \cite{AGeneralFrameworkForComparingPredictionsAndME}, expressing the magnitude of the associations between an independent and a dependent feature, having an intuitive interpretation of its value \cite{MEQuantifyingTheEffectOfChanges}.

Although not a new technique, the Partial Effect could be more explored in symbolic regression as a model-specific explanatory method since it is possible to automatically differentiate symbolic regression models, assuming all the functions from the primitive set are differentiable. In \cite{MeasuringFeatureImportanceOfSymbolicRegressionModels} the authors proposed the usage of Partial Effects as a model-specific explainer.

\begin{definition}[Partial Effects -- Local Feature Explanation]
    The local Partial Effect importance for a given observation $\mathbf{x}$ is given by the gradient of the model evaluated at $\mathbf{x}$:
    
    \begin{equation}
            \psi_j(\widehat{f}, \mathbf{x}) = \frac{\partial}{\partial X_j} \widehat{f}(\mathbf{x}),
    \end{equation}
    
    \noindent where $\widehat{f}$ is a regression analysis model.
\end{definition}

Similar to SHAP, we can calculate the Global explanations as the average of the absolute importance. The global importance of a feature $X_s$ is made by marginalizing its co-features to a representative value and evaluating the mean of all partial derivatives with values that $X_s$ can assume in the given data.

\begin{definition}[Partial Effects Global Feature Explanation]
    
    The global explanation for a feature $j$ is obtained by marginalizing all co-features and taking the mean absolute value of the local partial effects:
    
    \begin{equation}
        \phi_j(\widehat{f}, \mathcal{X}) = \frac{1}{n} \sum_{i=1}^{n} \bigg\lvert \frac{\partial}{\partial X_j} \widehat{f}(\mathcal{X}_{i, j}, \mathcal{X}_{i, C}) \bigg\rvert,
    \end{equation}
    
    \noindent where $\mathcal{X}_{i, C}$ is the co-features marginalized.
\end{definition}

\section{Measuring explanations quality}~\label{sec:measuringexplanationsquality}

Measuring the quality of the explanations is an open question \cite{CanITrustTheExplainer,TowardsQuantitativeEvaluationOfInterpretabilityMethods, TheStrugglesofFeatureBasedExplanations} as the definition of interpretability lacks mathematical rigor \cite{InterpretabilityAndExplainabilityAMLZoo}.
One possibility is to include humans in the loop to measure their perception of the quality of the explanations. Nevertheless, the results can be prone to biases due to the experimental design, leading the experimenters to consider multiple additional factors. Also, working with humans can limit the scalability of the framework, compromising the number of results needed for large-scale performance validation. Including humans in the loop of development and deployment of ML systems also requires proper ways of inspecting and analyzing the models, thus needing better benchmarks to validate the proposed explanatory methods.

In the literature, there are few measures proposed to evaluate the quality of the explanations focused on their robustness~\cite{RegularizingBlackBoxModelsForImprovedInterpretability, TowardsRobustInterpretability, OnTheInfidelityAndSensitivityOfExplanations, SLIME}. These measures are based on the principle that a high-quality explanation should not have noticeable variation caused by subtle perturbations in the explained point.

Another possibility explored in this work is to measure the quality of the explanation given an imperfect regression model. For this purpose, we can apply the explainers to a ground-truth model and compare how much the explanations differ when applied to different regression models. The following subsections will explain each of the measures considered in this work.

\subsection{Robustness of explanations}

Robustness measures evaluate explainers in how subtle perturbations in the observations affect the explanations and are suited to evaluate local explainers only. It is based on the \textit{desiderata} that a good-quality explanation should not vary with a small perturbation on the explained observation.

\subsubsection{Stability}

The stability measure~\cite{TowardsRobustInterpretability} is the degree to which the local explanation changes for a given point compared to its neighbors. A significant value implies that when the feature being explained changes in a small proportion, the feature importance responds in a more significant proportion, indicating that the explainer is not reliable since it is not locally stable.

\begin{definition}[Stability measure]
    The stability of an explanatory model for an observation $\mathbf{x}$ is calculated by:
    
    \begin{equation}
        \textup{S}(\widehat{f}, \psi, \mathbf{x}) = \mathbb{E}_{\mathbf{x}'\sim N_{\mathbf{x}}} \left [ \lvert\lvert \psi(\widehat{f}, \mathbf{x}) - \psi(\widehat{f}, \mathbf{x}')  \lvert\lvert_2^2 \right ],
    \end{equation}
    
    \noindent with $N_{\mathbf{x}}$ being the neighborhood of $\mathbf{x}$. 
\end{definition}

The stability evaluates the mean distance between the explanation for the original observation and for the explanation of all sampled neighbors. In \cite{RegularizingBlackBoxModelsForImprovedInterpretability} the authors uses a normal distribution to generate the neighborhood $N_{\mathbf{x}} = \mathcal{N}(\mathbf{x}, \sigma)$ with a previously defined value for $\sigma$.

The problem when using a normal distribution is that the neighborhood can have a different distribution from the training data, especially in a high-dimensional case.
As such, we propose two modifications for this measure:

\begin{enumerate}
    \item[1.] The neighborhood should be generated using a multivariate normal distribution $\mathcal{N}_d(\mathbf{\mu}, \mathbf{\Sigma})$. This should make the neighborhood more representative to the training data. This requires a variance-covariance matrix $\Sigma$, estimated using the training data. 
    \item[2.] The neighborhood spread should use the mean vector of each feature, multiplied by a neighborhood-range parameter $\lambda$ as the value of $\sigma$ since a fixed value is too conservative.
\end{enumerate}

To summarize, we propose the neighborhood to be calculated by: 

\begin{equation}
    N_{\mathbf{x}} = \mathcal{N}_d(\mathbf{x}, \lambda \cdot \textup{cov}(\mathcal{X})),
\end{equation}

\noindent where $\mathcal{X}$ is the training matrix where each line is an observation and each column is a feature, $\mathbf{x} \in \mathbb{R}^d$ is the observation of interest, and $\textup{cov}(\mathcal{X})$ is the variance-covariance matrix evaluated over the training data. The parameter $\lambda$ controls the size of the neighborhood and can be found experimentally, set to a fixed value, or calculated over the training data.

\subsubsection{(in)fidelity}

Taking the subset of most relevant features for an observation $\mathbf{x}$, it is expected that the explanation will attribute high values for those features rather than the others when subtle changes occur in $\mathbf{x}$. The idea of infidelity \cite{OnTheInfidelityAndSensitivityOfExplanations} is to measure the difference between two terms:
 
\begin{enumerate}
    \item[i.] The dot product between a significant perturbation $\mathbf{p} \in \mathbb{R}^{d}$ to a given observation $\mathbf{x}$ we are trying to explain and its corresponding explanation, and
    \item[ii.] The difference in the prediction between the perturbed and original observations.
\end{enumerate} 

\begin{definition}[Infidelity measure]

    \begin{equation}
        \textup{INFD}(\widehat{f}, \psi, \mathbf{x}) = \mathbb{E}_{\mathbf{p} \in \mu_\mathbf{p} } \left [ \left  ( \underbrace{\mathbf{p}^T \psi(\widehat{f}, \mathbf{x})}_{\textit{i}} - \underbrace{(\widehat{f}(\mathbf{x}) - \widehat{f}(\mathbf{x} - \mathbf{p}))}_{\textit{ii}} \right )^{2} \right ]
    \end{equation}
    
    \noindent where $\psi$ is an importance attribution explainer, $\widehat{f}$ is a black-box model, and $\mathbf{p}$ is a random feature representing perturbations around the point of interest $\mathbf{x}$ generated by a probability measure $\mu_{\mathbf{p}}$.
\end{definition}

A good explanation should be robust to changes in the prediction in response to significant perturbations. This measure is helpful to evaluate if the explanation is robust to miss-specifications or noise in the given observation.

A significant value of infidelity means that the explainer changes more abruptly when small changes in the observation occur, resulting in very different feature importance vector, and thus is not reliable.
In this paper, we sample $\mathbf{p}$ with $\mu_\mathbf{p} \sim \mathbf{x} - \mathcal{N}_d(\mathbf{x}, \lambda \cdot \textup{cov}(\mathcal{X}))$.

\subsubsection{Jaccard Stability}

The Jaccard Index was used in \cite{SLIME} to measure the stability of their proposed explainer called S-LIME. The Jaccard Index is often used as a similarity metric between two sets, and it is defined as the size of the intersection between the sets by the size of the union of these sets. If they are both equal, the Jaccard Index will have a value of $1$. To use this metric for stability, we verify whether the set of the $k$ most important features, measured by the magnitude of their importance, changes within a neighborhood.

\begin{definition}[Jaccard Stability]

    Let $\mathbf{x}$ be a observation being explained, and let $\tilde{\psi}(\mathbf{x})_{k}$ denote the subset of $k$ most important features for $\mathbf{x}$. The Jaccard Stability measure is calculated by:

    \begin{equation}
        \textup{J}(\widehat{f}, \psi, \mathbf{x}) = \mathbb{E}_{\mathbf{x}'\sim N_{\mathbf{x}}} \left [ \frac{\lvert \tilde{\psi}(\mathbf{x})_{k} \cap \tilde{\psi}(\mathbf{x}')_{k} \lvert}{\lvert \tilde{\psi}(\mathbf{x})_{k} \cup \tilde{\psi}(\mathbf{x}')_{k}\lvert} \right ],
    \end{equation}
    
    \noindent with $N_{\mathbf{x}}$ being the neighborhood of $\mathbf{x}$, and $k$ being the size of the subset. 
\end{definition}

\subsection{Quality of explanations with imperfect predictions}

Different explanatory methods have different interpretations of the importance values. For example, a Shapley value $\psi_i$ of a feature means how much it contributed to the difference in prediction from the average prediction. On the other hand, for Partial Effects, it means how the prediction will change if we make a slight variation to the input value. We cannot compare two explanatory methods directly since there is no guarantee of correspondence between explanations.

Nevertheless, since explanations are generated from regression models that are usually an imperfect approximation of the true model, we can measure how much information is lost when trying to explain these imperfect models. For this purpose, we assume the existence of a ground-truth explanation that we can use to compare the \textit{i)} error in direction (whether the importance is negative or positive) and the \textit{ii)} magnitude of the explanation extracted from the regression models; given that we have access to the ground-truth function that generated the data set.

When used to measure global quality, only one explanation vector is compared. When measuring local quality, each observation in the test data is evaluated with the quality measureand the average quality is reported.

\subsubsection{Cosine Similarity}

The cosine similarity, first used in \cite{EvaluatingLocalExplanationMethodsOnGroundTruth} to measure the quality of an explanation, returns a value between $-1$ and $1$ representing if the explanations are in opposed or same directions. A cosine similarity of $0$ means that the explanations are orthogonal.

\begin{definition}[Cosine Similarity between two explanations]~\label{def:cosinesim}
    Let $\overrightarrow{\phi}$ be a vector of feature importance generated by a global explanatory method $\phi$ (or, similarly, let $\overrightarrow{\psi}$ be a vector of local importance for an observation $\mathbf{x}$ generated by a local explanatory method) for a regression model $\widehat{f}$, and let $\overrightarrow{e}$ be the correct explanation generated from applying one of the explanation methods to the ground-truth model. The cosine similarity is calculated by:
    
    \begin{equation}
    \textup{cos}(\overrightarrow{e}, \overrightarrow{\psi}) = \frac{\overrightarrow{e} \cdot \overrightarrow{\psi}}{\|\overrightarrow{e}\| \|\overrightarrow{\psi}\|}.
    \end{equation}
\end{definition}

\subsection{Normalized Mean Squared Error}

The Normalized Mean Squared Error measures the squared difference between the ground-truth explanation and the evaluated explanation normalized by the true explanation variance. This captures whether the explanations differ not only in direction but also in magnitude.

\begin{definition}[Normalized Mean Squared Error]~\label{def:nmseexplicacao}
     Let $\overrightarrow{\phi}$ be a feature importance vector generated by a global explanatory method $\phi$ (or, similarly, let $\overrightarrow{\psi}$ be a vector of local importance for an observation $\mathbf{x}$ generated by a local explanatory method) for a regression model $\widehat{f}: \mathbb{R}^d \rightarrow \mathbb{R}$, and let $\overrightarrow{e}$ be an expected explanation (ground-truth). The Normalized Mean Squared Error (NMSE) is the MSE divided by the variance of $\overrightarrow{e}$:

    \begin{equation}
        \textup{NMSE}(\overrightarrow{e}, \overrightarrow{\phi}) = \frac{ \sum_{i=1}^{d}(e_i - \phi_i)^2}{\sum_{i=1}^{d}(e_i - \overline{e})^2},
    \end{equation}
    
    \noindent where $\overline{e}$ is the average of the feature importance values in $\overrightarrow{e}$.
\end{definition}

\section{Methodology}~\label{sec:methods}

\begin{figure}[t!]
    \centering
    \includegraphics[width=\linewidth]{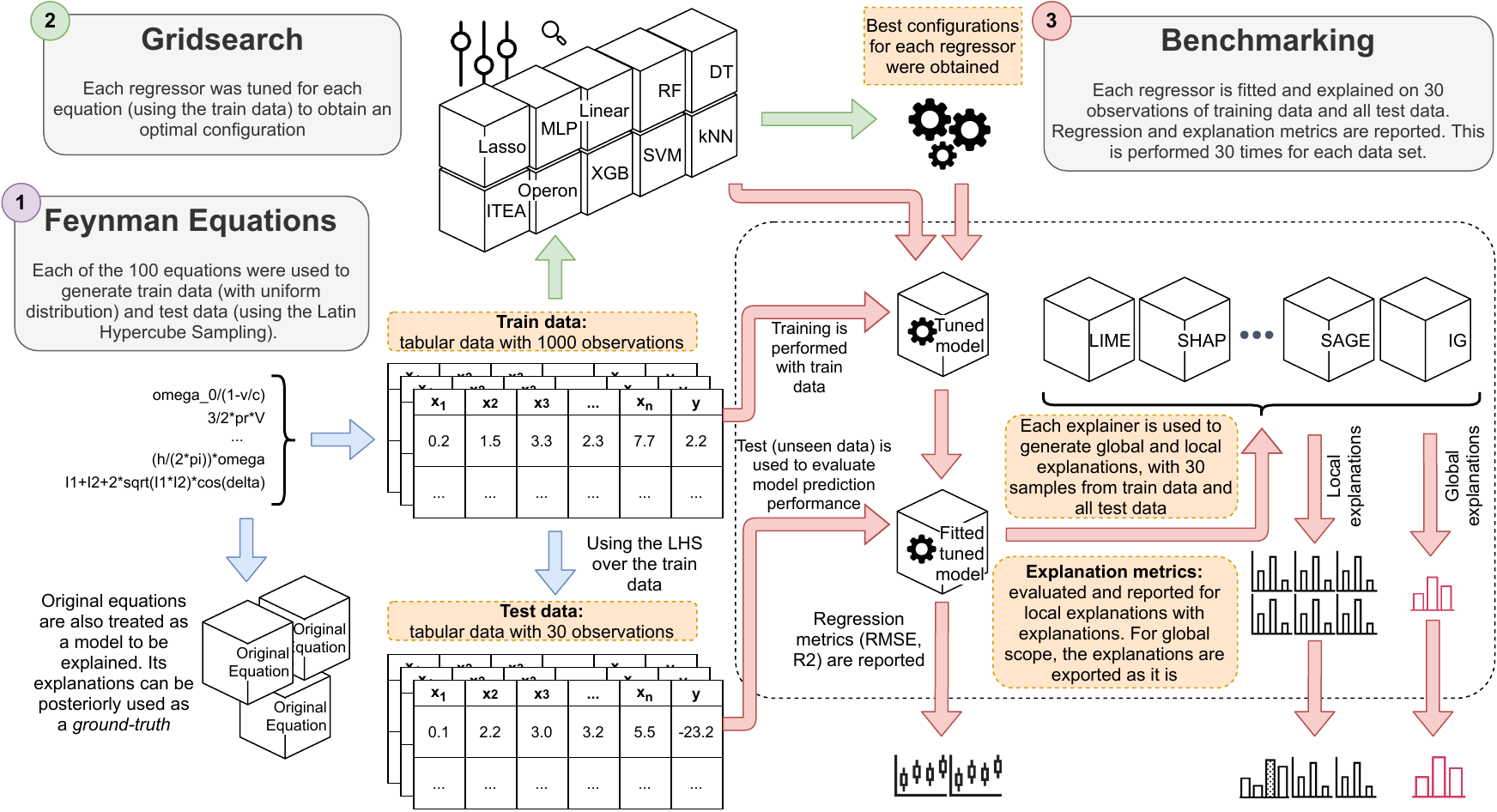}
    \caption{Scheme of the whole benchmark process to generate the results. The Feynman equations are used to generate the train and test data. Then the regression methods are fine-adjusted through a gridsearch process to finally be trained and used as input to feature importance methods.}
    \label{fig:benchmark-scheme}
\end{figure}

The experiments will follow the methodology illustrated in Fig. \ref{fig:benchmark-scheme}. In short, we will follow three steps.

First, we use the generating functions to create: \textit{i)} a predictor using the original equation as the prediction function, \textit{ii)} a training data set with $1000$ observations, and \textit{iii)} a test data set with 30 observations generated using the Latin Hypercube Sampling~\cite{loh1996latin} method based on the training data. Both train and test data do not contain any noise to prevent adding artifacts to the results.

The second step uses the training data to tune each regressor with a grid-search using $3$-fold cross-validation for each data set. The grid-search optimizes the configuration that maximizes the $R^2$ score. 

Finally, the third step executes the benchmark experiments using the optimal configurations obtained for each regression method. The models are generated using the training data and explained using different explanatory methods. Whenever the regression algorithm is stochastic, we repeat this procedure $30$ times for every data set.

We will analyze the results by measuring the prediction error of each regression model, followed by the quality measures of the explainer models. For local explanations, we will use the robustness measure and the quality measure and, for global explanations, we will only report the quality measure as the robustness measures require a local neighborhood. We will report heatmaps of the results for different combinations of regression and explanatory models.

The neighborhood-range factor used for the local explanations was $0.001$, and the number of top features for the Jaccard Stability was $k=1$.

\subsection{Data sets}

For the overall results, we use the Feynman benchmark, introduced in \cite{AIFeynman}, containing $100$ equations from mechanical, electromagnetism, and quantum physics, with different degrees of complexity. The reason for choosing this benchmark is that it contains problems that describe actual phenomena, where the problem features have physical or mathematical meaning. The original data contained $1,000,000$ observations, from which we randomly picked $1,000$ data points. The test set was generated using the Latin Hypercube.

For a more detailed analysis, we selected four benchmark problems commonly used in GP literature described in Table~\ref{tab:gpbench}.

\begin{figure}[t!]
    \centering
    \includegraphics[width=0.75\linewidth]{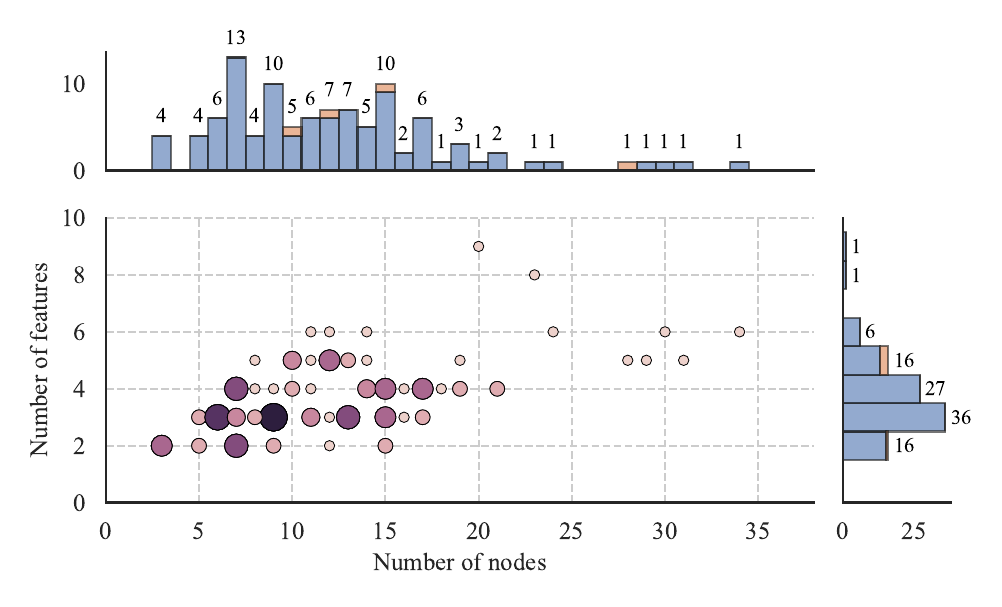}
    \caption{Joint distribution of the number of features and the number of nodes required to build the expression tree for the iirsBenchmark data sets. The lighter portion of the histograms represents the GP benchmark.}
    \label{fig:numofvariables}
\end{figure}

Fig. \ref{fig:numofvariables} shows the distribution of the number of features for these 104 data sets.

\begin{table}[t!]
\caption{Description of the additional 4 GP benchmark problems. $U(l_b, u_b, n_s)$ denotes a uniform distribution with bounds $[l_b, u_b]$ and $n_s$ observations, and $E(start, stop, step)$ is an evenly spaced grid in the interval $[start, stop]$ and step size given by $step$. We adapted the number of generated observations in relation to the original publication to make the results and comparisons compatible with the Feynman benchmark.}
\centering
\label{tab:gpbench}
\begin{tabular}{@{}lllll@{}}
\toprule
Name & Formula & Features & Train set/Test set \\ \midrule
Korns-11 & $6.87 + 11 \textup{cos}(7.23 x^3)$ & $\{x, y, z, v, w\}$ & \begin{tabular}[t]{l}$U(-50, 10, 1000)$\\ $U(-50, 10, 100)$ \\\end{tabular} \\

Korns-12 & $2-2.1\textup{cos}(9.8x)\textup{sin}(1.3w)$ & $\{x, y, z, v, w\}$ & \begin{tabular}[t]{l}$U(-50, 10, 1000)$\\ $U(-50, 10, 100)$ \\\end{tabular} \\

Vladislavleva-4 & $\frac{10}{5+\sum_{i=1}^{5}(x_i - 3)^2}$ & $\{x_i\}_{i=1}^{5}$ & \begin{tabular}[t]{l}$U(0.05, 6.0, 1000)$\\ $U(-0.5, 10, 100)$ \\\end{tabular} \\

Pagie-1 & $\frac{1}{1+x^{-4}}+\frac{1}{1+y^{-4}}$ & $\{x, y\}$ & \begin{tabular}[t]{l}$E(-5, 5, 0.01)$\\ $E(-5, 5, 0.1)$ \\\end{tabular} \\\bottomrule
\end{tabular}
\end{table}

\subsection{Regression and explanatory methods}

Table \ref{tab:regressorsiirs} shows all the regression algorithms used in this experiment. All of them, except for ITEA\footnote{\url{https://github.com/gAldeia/itea-python}.} and Operon\footnote{\url{https://github.com/heal-research/operon}.}, are available via scikit-learn module\footnote{\url{https://github.com/scikit-learn/scikit-learn}.}.

\begin{table}[t!]
    \centering
    \caption{Regression methods available on \textit{iirsBenchmark} and used in the experiments.}
    \label{tab:regressorsiirs}
    \begin{tabular}{@{}lll@{}}
    \toprule
        Regressor & Interpretability & Method type \\ \midrule
        XGB & black-box & Tree boosting\\
        RF & black-box & Tree bagging \\
        MLP & black-box & Feed-forward Neural network  \\
        SVM & black-box & Vector machine  \\
        k-NN & gray-box & Instance method \\
        Operon & gray-box & Symbolic regression  \\
        ITEA & gray-box & Symbolic regression  \\
        Linear regression & white-box & Regression modeling  \\
        LASSO regression & white-box & Regression modeling  \\
        Single decision tree & white-box & Decision tree \\
        \botrule
    \end{tabular}
\end{table}

The tested configurations for each regressor are reported in Table \ref{tab:regressors}. For symbolic regression methods ITEA and Operon, the functions sets were, respectively, $\{\textup{log, sqrt, id, sin, cos, tanh, exp, expn, arcsin}\}$ and $\{+, -, \times, /, \textup{exp, log, sqrt, square, sin, cos, tanh, asin, constant, variable}\}$.

\begin{table}[t!]
    \caption{Regressors considered in the experiments, with their hyper-parameters to be adjusted through the gridsearch procedure. The Linear regressor is the only that had no adjustment.}
    \label{tab:regressors}
    \centering
    \begin{tabular}{@{}lrl@{}}
    \toprule
    Regressor & Hyper-parameter & Gridsearch values \\ \midrule
    XGB & \begin{tabular}[t]{@{}r@{}}n\_estimators\\ min\_samples\_split\end{tabular} & \begin{tabular}[t]{@{}l@{}}$[100, 200, 300]$\\ $[0.01, 0.05, 0.1]$\end{tabular} \\
    RF & \begin{tabular}[t]{@{}r@{}}n\_estimators   \\ min\_samples\_split\end{tabular} & \begin{tabular}[t]{@{}l@{}}$[100, 200, 300]$\\ $[0.01, 0.05, 0.1]$\end{tabular} \\
    MLP & \begin{tabular}[t]{@{}r@{}}hidden\_layer\_sizes\\ activation\end{tabular} & \begin{tabular}[t]{@{}l@{}}$[(50,), (50, 100, ), (100,), (100, 100, )]$\\ $['identity', 'logistic', 'tanh', 'relu']$\end{tabular} \\
    SVM & \begin{tabular}[t]{@{}r@{}}kernel\\ degree\end{tabular} & \begin{tabular}[t]{@{}l@{}}$['linear', 'rbf', 'poly']$\\ $[1, 2, 3, 4]$\end{tabular} \\
    k-NN & n\_neighbors & $[3, 5, 7, 9, 11, 17, 19, 23, 29, 31]$ \\
    Operon & \begin{tabular}[t]{@{}r@{}}population\_size\\ generations\end{tabular} & \begin{tabular}[t]{@{}l@{}}$[100, 250, 500]$\\ $[100, 250, 500]$\end{tabular} \\ 
    ITEA & \begin{tabular}[t]{@{}r@{}}popsize\\ gens\end{tabular} & \begin{tabular}[t]{@{}l@{}}$[100, 250, 500]$\\ $[100, 250, 500]$\end{tabular} \\
    LASSO & alpha & $[0.001, 0.01, 0.1, 1, 10]$ \\
    Decision Tree & \begin{tabular}[t]{@{}r@{}}max\_depth     \\ max\_leaf\_nodes\end{tabular} & \begin{tabular}[t]{@{}l@{}}$[5, 10, 15]$\\ $[5, 10, 15]$\end{tabular} \\
    \botrule
    \end{tabular}
    
    \footnotetext{Default values of hyper-parameters were omitted. All the regressors have their respective python module, and omitted parameters default values can be found on each regressor documentation}
\end{table}

Table \ref{tab:explainers} shows the explanatory methods used in this experiment together with their main characteristics. We also created a \textit{Random Importance} explainer that supports local and global explanations, attributing random ranks of importance for each feature every time it is called. This explainer will serve as a baseline as it is expected to be the worst explainer. 

For the robustness measures, we also report the values obtained in the ground-truth model, called \emph{Feynman explainer}. This is the result of applying the explainer models to the ground-truth expression, representing the robustness of the true model. 

\begin{table}[t!]
    \centering
    \caption{Explanation methods available in \textit{iirsBenchmark} and used in the experiments.}
    \label{tab:explainers}
    \begin{tabular}{@{}lllll@{}}
    \toprule
        Explainer & Agnostic & Local & Global & Method type \\ \midrule
        Permutation Importance & \textbf{Y} & N & \textbf{Y} & feature removal \\
        SHAP & \textbf{Y} & \textbf{Y} & \textbf{Y} & feature removal \\
        SAGE & \textbf{Y} & N & \textbf{Y} & feature removal \\
        LIME & \textbf{Y} & \textbf{Y} & N & Approximate by linear model \\
        ELA & N & \textbf{Y} & N & Approximate by linear model \\
        Morris sensitivity & \textbf{Y} & N & \textbf{Y} & Sensitivity analysis \\
        Integrated Gradients & \textbf{Y} & \textbf{Y} & N & Gradient analysis \\
        Partial Effects (PE) & N & \textbf{Y} & \textbf{Y} & Gradient analysis \\
        Random Importance & \textbf{Y} & \textbf{Y} & \textbf{Y} & Random attribution \\
    \botrule
    \end{tabular}
\end{table}

\subsection{Result analysis}

The results involving multiple data sets are reported by presenting the median and the Interquartile Range (IQR). While boxplots implicitly report the Interquartile Range (IQR) --- being the size of the filled box in the graphs ---, tables and heatmaps reports the results as $\tilde{m} \pm \textup{IQR}$, where $\tilde{m}$ is the median of the group and $\textup{IQR}$ is the respective IQR. Those descriptors were chosen due to robustness to outliers while still providing a good summary of the distribution.

We use \textit{Critical Diagrams} \cite{StatisticalComparisonsOfClassifiers} to represent the acceptance or rejection of the null hypothesis. The critical diagrams sorts the compared groups by their average ranks, and two or more methods are connected by a horizontal line if there is no statistical significance between them. The statistical test used was the Wilcoxon signed-rank, with the $p$-values corrected by the Holm-Bonferroni method, recommended when testing only one null hypothesis that all groups have no significant differences between themselves \cite{Whatistheproperwaytoapplymultiplecomparison}. The correction method mitigates the alpha inflation when multiple comparisons are made. The statistical significance threshold considered is $p < 0.05$, evaluated after correction.

Finally, since the computational cost of the explanation may be relevant to some applications, we also report the runtime of each explainer for both local and global explanations.

\subsection{\textit{iirsBenchmark}}

All regressors and explanatory methods included in the experiments were wrapped up in a package module with a syntax similar to the Python ML library scikit-learn in the \textit{iirsBenchmark}. The python module also provides a script to execute the experiments and all post-processing scripts.

All regression methods were previously implemented either by the scikit-learn module or by their corresponding authors following scikit guidelines, all of which have a structured unification. The \textit{iirsBenchmark} only extends those regressors by adding more class methods and static attributes useful in the interpretability context. The explainers being evaluated lack a common interface. In spite of that, we unified their interface by implementing standardized classes in \textit{iirsBenchmark}.

\section{Experimental Results}~\label{sec:results}

In the first part of this section, we report and analyze the results obtained with the Feynman benchmark. We present the aggregated results of all combinations of regression models and explanation models. The following subsection shows the results obtained on the four select challenging benchmarks from SR literature, in which we will make a more detailed analysis of the obtained results.

\subsection{Feynman benchmark}~\label{sec:feymanresults}

In this subsection, we will have a broader view of the results for a benchmark set that does not present a challenging scenario for the non-linear regression models. We expect to observe explanations close to what we would obtain using the generating function.

\subsubsection{Model accuracy}~\label{sec:modelacc}

\begin{figure}[t!]
    \centering
    
    \begin{subfigure}[b]{0.5\textwidth}
        \centering
        \includegraphics[width=\linewidth]{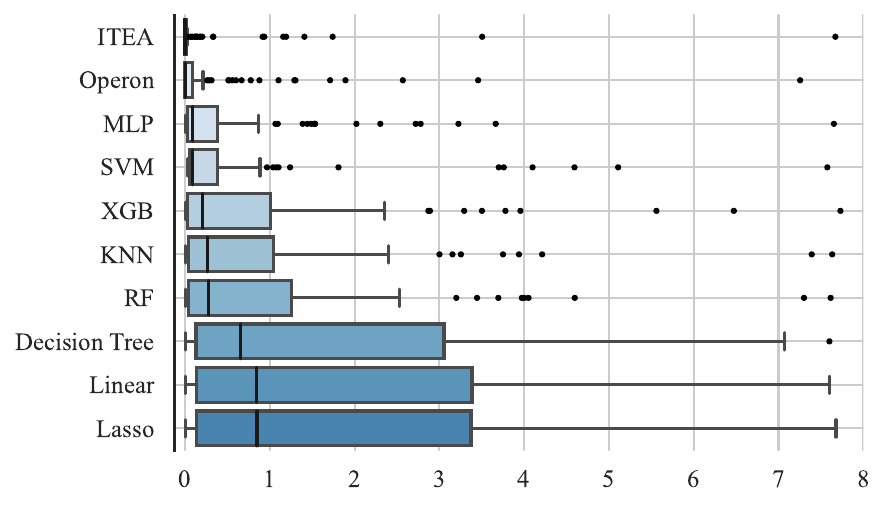}
        \caption{MAE on test}
        \label{fig:r2_test_boxplot}
    \end{subfigure}%
    \hfill
    \begin{subfigure}[b]{0.5\textwidth}
        \centering
        \includegraphics[width=\linewidth]{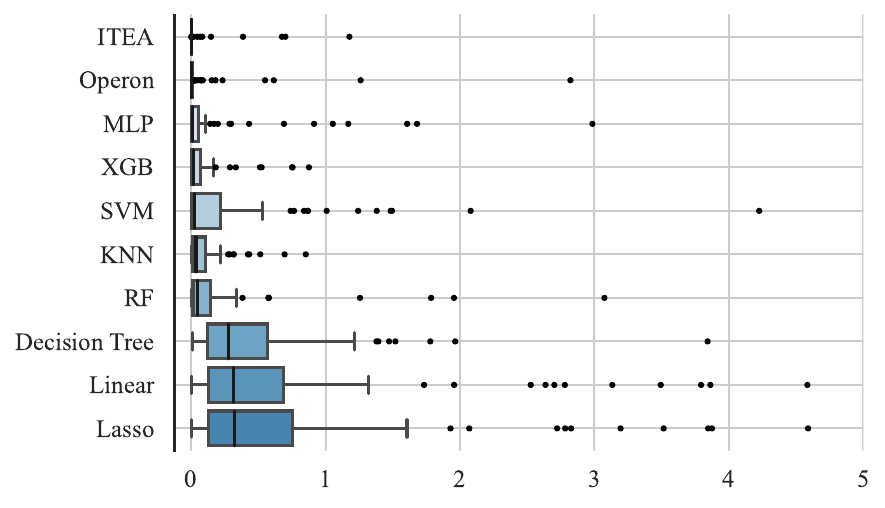}
        \caption{NMSE on test}
        \label{fig:nmse_test_boxplot}
    \end{subfigure} \\%
    \begin{subfigure}[b]{0.5\textwidth}
        \centering
        \includegraphics[width=\linewidth]{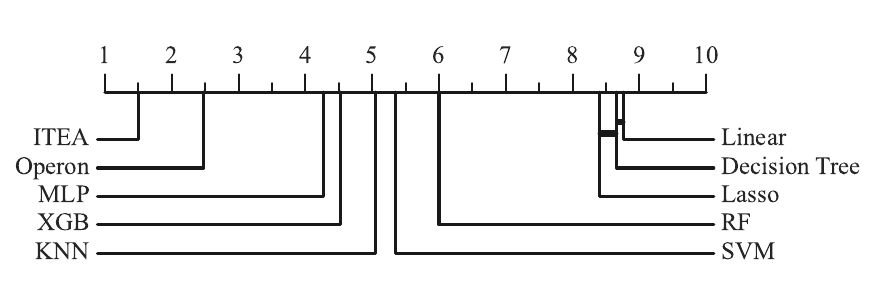}
        \caption{Critical Diagram for MAE on test}
        \label{fig:r2_test_cd}
    \end{subfigure}%
    \hfill
    \begin{subfigure}[b]{0.5\textwidth}
        \centering
        \includegraphics[width=\linewidth]{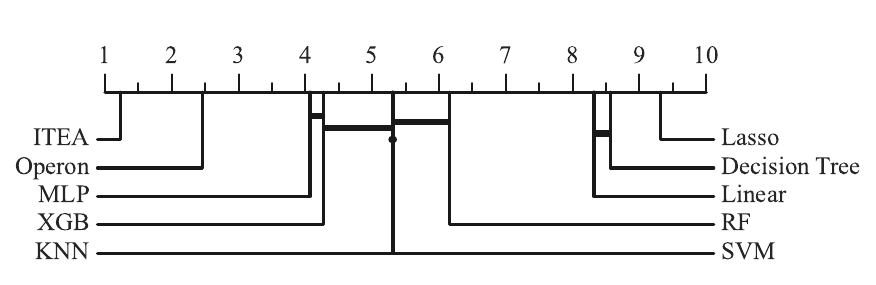}
        \caption{Critical Diagram for NMSE on test}
        \label{fig:nmse_test_cd}
    \end{subfigure} \\%
    
    \caption{MAE (smaller is better) and NMSE (smaller is better) boxplots for all regressors, vertically ordered from the best to worst median values. The Critical Diagram below each plot indicates the absence of statistical significance between groups connected by a horizontal bar.}
    \label{fig:nmse_r2_test_regression}
\end{figure}

Fig.~\ref{fig:nmse_r2_test_regression} shows the boxplot of the MAE and NMSE of each method considering the entirety of the Feynman benchmark, with the critical difference diagram of the average rank for each method. As we can see from these plots, ITEA and Operon found the best performing regression models in the benchmark. The critical difference diagrams show a significant difference between ITEA and Operon, even though both obtained models are close to the maximum observed accuracy. The three models considered interpretable in the literature were the worst-performing models in this benchmark, corroborating with the idea of the trade-off between interpretability and accuracy.

\begin{figure}[t!]
    \centering
    \begin{subfigure}[b]{0.5\textwidth}
        \centering
        \includegraphics[width=\linewidth]{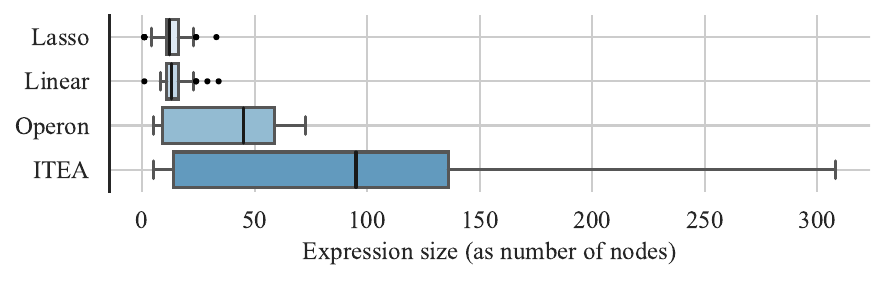}
        \caption{Expression sizes}
        \label{fig:expression_sizes_boxplot}
    \end{subfigure}%
    \begin{subfigure}[b]{0.5\textwidth}
        \centering
        \includegraphics[width=\linewidth]{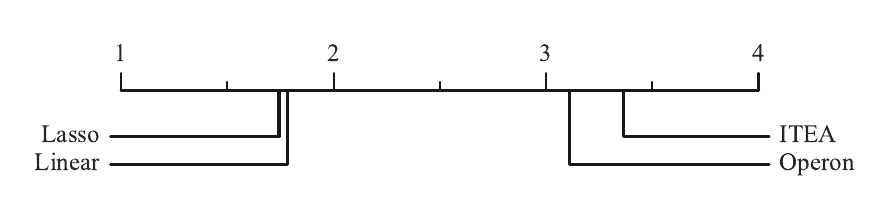}
        \caption{Critical Diagram for expression sizes}
        \label{fig:expression_sizes_cd}
    \end{subfigure} \\%
    \caption{Expression sizes as number of nodes (smaller is better) for regression models. The regression methods are vertically ordered from the best to worst.}
    \label{fig:expression_sizes}
\end{figure}

Another relevant aspect of the interpretability of the model is the expression size. Larger expressions are harder to read and understand. Fig.~\ref{fig:expression_sizes} shows the boxplot of the distribution of expression sizes obtained by Lasso, Linear, Operon, and ITEA. Lasso and Linear obtained the smaller expressions since their size is bounded by the number of features, followed by Operon with a median size of $50$. ITEA median size was $100$ with an upper quartile much higher than Operon. This means that, despite the difference in accuracy between ITEA and Operon, Operon managed to find smaller expressions than ITEA, possibly increasing its readability by human experts.

\begin{figure}[t!]
    \centering
    \begin{subfigure}[b]{0.475\textwidth}
        \centering
        \includegraphics[width=\linewidth]{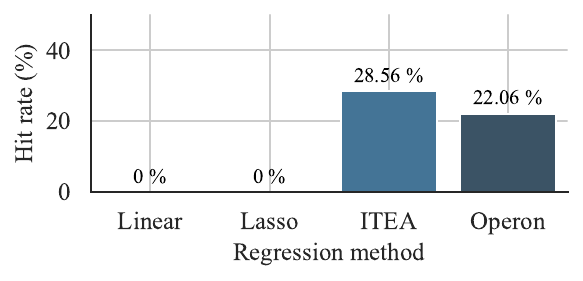}
        \caption{Hit rate percentage on Feynman data sets}
        \label{fig:hit_rate_boxplot}
    \end{subfigure}%
    \hfill
    \begin{subfigure}[b]{0.475\textwidth}
        \centering
        \includegraphics[width=0.8\linewidth, trim={0 1.5cm 0 1.5cm}, clip=true]{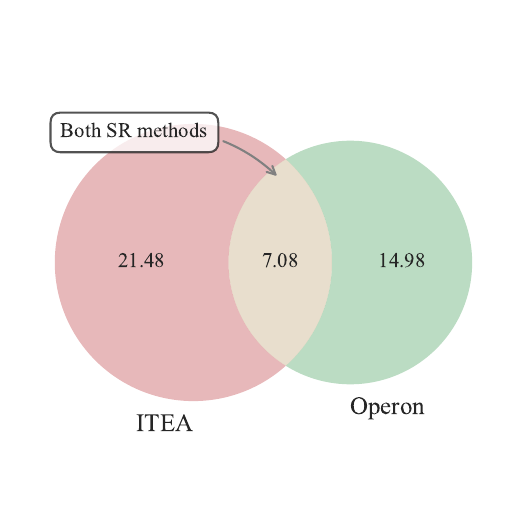}
        \caption{Venn diagram of hit rate percentage for the two symbolic regression methods}
        \label{fig:hit_rate_venn}
    \end{subfigure} \\%
    
    \caption{Hit rate on Feynman data sets calculated for regression methods that return mathematical expressions.}
    \label{fig:hit_rate_results}
\end{figure}

Complementing the analysis of the expression size, Fig.~\ref{fig:hit_rate_boxplot} shows a bar plot of the hit rate of each of these methods. Given the original expression $f$ and the SR model $\widehat{f}$, we perform a symbolic simplification of the expression $\widehat{f} - f$, considering a hit if the result of the simplification is zero. The hit rate is the percentage of expressions obtained by each model that corresponds precisely to the ground-truth generating function.

As expected, ITEA obtained a higher hit rate than Operon since almost half of this benchmark is representable by the constrained representation of ITEA, having a search space that tends to favor this SR method. Fig.~\ref{fig:hit_rate_venn} shows a Venn diagram of the intersection of the data sets that both methods obtained the generating function. Only $7\%$ were mutually found by both methods, meaning that by merging the results of both methods, SR obtained a hit rate of about $42\%$ on this benchmark. These results highlight the possibility of using SR as an innate interpretable model without using a \emph{post-hoc} explanatory method.

\subsubsection{Explanation computational cost}

Table~\ref{tab:explanation_execution_time} shows the median and IQR of the execution time for each explainer when combined with different models. Regarding the global methods, Partial Effect, Permutation Importance, and Morris Sensitivity have a significantly smaller computation run-time than SHAP and SAGE. This is expected since the former methods only require the evaluation of a well-defined function to every point in the data set, while the latter methods require the sampling of different subsets of features. We should notice that it is possible to reduce the run-time of both SHAP and SAGE with the cost of reducing the explanation quality.

For the local explanations, Partial Effects are the fastest method since it becomes just a function applied to a given instance after calculating the symbolic derivative. The other methods are comparable regarding run-time but with some combinations that stand out: SHAP and IG with Random Forest and IG with ITEA require a significantly longer time when compared to other regressors using the same explanation method. These methods require a higher number of evaluations to generate the feature importance, and so they are more affected by prediction model size. Specifically for SR methods, we used the model agnostic version of IG that approximate derivatives to a fair performance comparison.

\begin{table}[t!]
    \centering
    \caption{Median and IQR of execution time (in seconds) each explainer took to generate the explanations.}
    \label{tab:explanation_execution_time}
    \footnotesize
    \begin{tabular}{@{}rc@{\extracolsep{7pt}}c@{\extracolsep{7pt}}c@{\extracolsep{7pt}}c@{\extracolsep{7pt}}c@{}}
        \toprule
       \begin{tabular}[c]{@{}c@{}}Global\\Explanation\end{tabular} & \begin{tabular}[c]{@{}l@{}}Partial\\ Effects\end{tabular} & SHAP & SAGE & \begin{tabular}[c]{@{}l@{}}Permutation\\ Importance\end{tabular} & \begin{tabular}[c]{@{}l@{}}Morris\\ Sensitivity\end{tabular} \\ \midrule
        
        KNN& ---& $ 21.07 \pm19.71$& $ 1225.18 \pm5217.46$& $0.70 \pm 0.82$& $0.30 \pm 0.31$\\
        Linear & $0.00 \pm 0.00$& $ 14.47 \pm14.91$& $ 27.21 \pm 110.25$& $0.06 \pm 0.06$& $0.36 \pm 0.38$\\
        Lasso& $0.00 \pm 0.00$& $ 14.05 \pm11.00$& $ 21.45 \pm 107.16$& $0.06 \pm 0.06$& $0.31 \pm 0.29$\\
        Decision Tree & ---& $ 14.30 \pm14.07$& $ 38.83 \pm 100.23$& $0.06 \pm 0.08$& $0.30 \pm 0.31$\\
        RF & ---& $111.42 \pm94.86$& $ 3116.55 \pm9144.32$& $6.94 \pm 6.23$& $0.42 \pm 0.34$\\
        MLP& ---& $ 40.21 \pm19.14$& $549.54 \pm3613.16$& $1.87 \pm 1.59$& $0.44 \pm 0.33$\\
        SVM& ---& $ 18.74 \pm26.52$& $ 2077.81 \pm 10092.26$& $1.28 \pm 3.55$& $0.33 \pm 0.32$\\
        XGB& ---& $ 18.14 \pm19.20$& $824.13 \pm3644.58$& $0.78 \pm 0.70$& $0.29 \pm 0.28$\\
        Operon & $0.66 \pm 0.46$& $ 12.37 \pm 9.77$& $ 25.57 \pm 159.81$& $0.06 \pm 0.04$& $0.21 \pm 0.09$\\
        ITEA & $0.14 \pm 0.13$& $ 25.41 \pm22.89$& $486.43 \pm1869.14$& $1.55 \pm 1.01$& $0.41 \pm 0.28$\\
        \midrule
        
        \begin{tabular}[c]{@{}c@{}}Local\\ Explanation\end{tabular} & \begin{tabular}[c]{@{}l@{}}Partial\\ Effects\end{tabular} & SHAP & \begin{tabular}[c]{@{}l@{}}Integrated\\ Gradients\end{tabular} & LIME & ELA \\ \midrule
        
        KNN& ---& $2.12 \pm 1.87$& $4.53 \pm 3.82$& $3.73 \pm 2.99$& ---\\
        Linear & $0.00 \pm 0.00$& $1.37 \pm 1.52$& $0.92 \pm 0.69$& $4.25 \pm 3.47$& $1.83 \pm 1.68$\\
        Lasso& $0.00 \pm 0.00$& $1.40 \pm 1.16$& $0.86 \pm 0.73$& $3.65 \pm 2.75$& $1.75 \pm 1.49$\\
        Decision Tree & ---& $1.26 \pm 1.21$& $0.94 \pm 0.84$& $3.70 \pm 3.19$& ---\\
        RF & ---& $ 11.41 \pm10.15$& $167.02 \pm 154.94$& $6.93 \pm 6.95$& ---\\
        MLP& ---& $3.98 \pm 1.71$& $ 11.41 \pm 8.51$& $6.52 \pm 4.76$& ---\\
        SVM& ---& $1.98 \pm 2.91$& $1.02 \pm 0.73$& $3.30 \pm 2.63$& ---\\
        XGB& ---& $1.70 \pm 1.86$& $2.36 \pm 2.24$& $3.76 \pm 3.69$& ---\\
        ITEA & $0.13 \pm 0.14$& $2.71 \pm 2.24$& $ 17.67 \pm16.99$& $4.02 \pm 2.68$& $2.16 \pm 1.71$\\
        Operon & $0.06 \pm 0.04$& $1.09 \pm 0.89$& $0.75 \pm 0.48$& $2.93 \pm 1.48$& $1.25 \pm 0.49$\\
        \bottomrule
    \end{tabular}
\end{table}

\subsubsection{Local Explanations}~\label{sec:localexpls}

\begin{figure}[t!]
    \begin{subfigure}[b]{\textwidth}
        \centering
        \includegraphics[width=\linewidth]{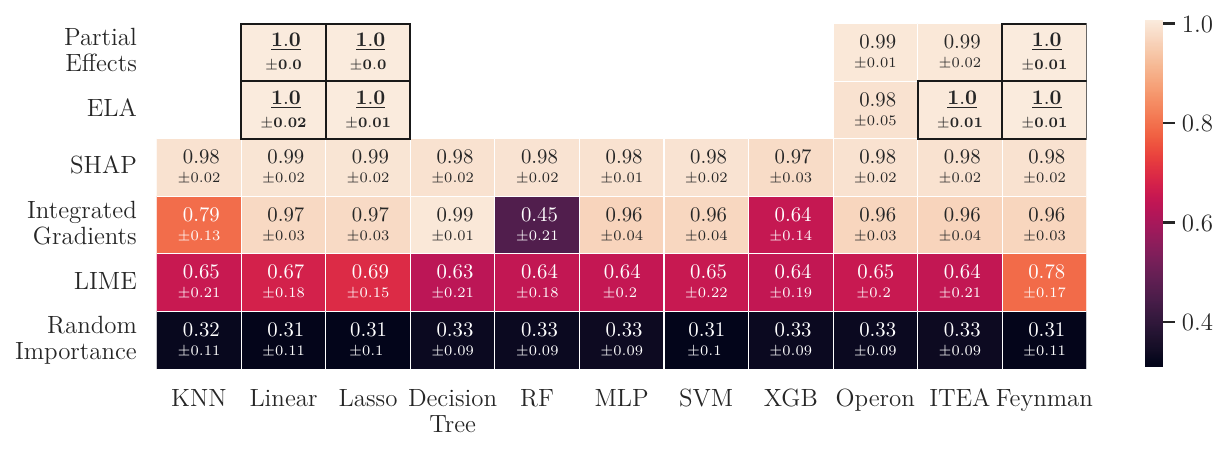}
        \caption{Jaccard Stability (greater is better) \textit{heatmap} on the Feynman data sets.}
        \label{fig:jaccard_heatmap}
    \end{subfigure} \\%
    \begin{subfigure}[b]{\textwidth}
        \centering
        \includegraphics[width=\linewidth]{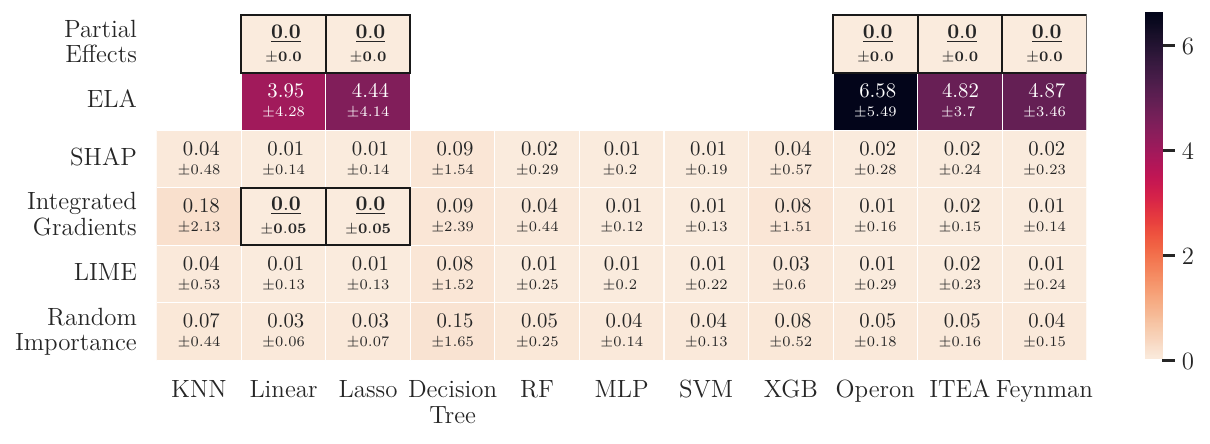}
        \caption{Infidelity (smaller is better) \textit{heatmap} on the Feynman data sets.}
        \label{fig:infidelity_heatmap}
    \end{subfigure} \\%
    \begin{subfigure}[b]{\textwidth}
        \centering
        \includegraphics[width=\linewidth]{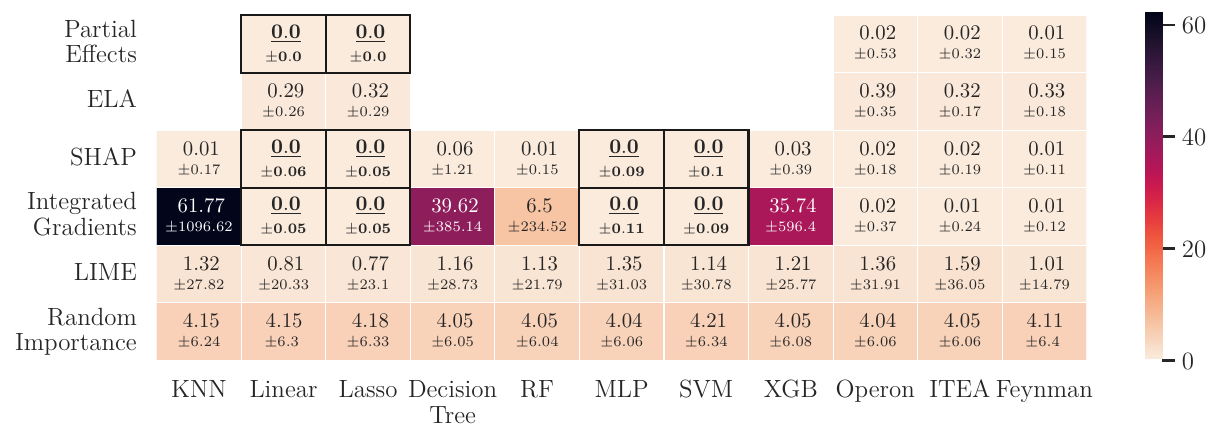}
        \caption{Stability (smaller is better) \textit{heatmap} on the Feynman data sets.}
        \label{fig:stability_heatmap}
    \end{subfigure}
    
    \caption{Local explanation robustness metrics \textit{heatmaps} showing the median and IQR for each explainer-regressor combination. The best values cells are highlighted with a black edge.}
    \label{fig:local_robustness_heatmaps}
\end{figure}

Fig.~\ref{fig:local_robustness_heatmaps} show the heatmap plot of different robustness measures for the combinations of regressors and explanation models for the local explanations. The Jaccard Stability (Fig.~\ref{fig:jaccard_heatmap}) is a very sensitive measure since we are calculating based only on the top feature, so either the explanation gets everything right or everything wrong in a single explanation. We can see that, despite this difficulty, Partial Effects, ELA and SHAP obtained the highest score for every prediction model. This means that the top feature does not change around a small neighborhood. Integrated Gradient presented stability issues under this measure for the ensemble methods and KNN. LIME also presented worse results for every regressor but still higher than the Random Importance.

Regarding Infidelity (Fig.~\ref{fig:infidelity_heatmap}), ELA was the only explanation method that obtained worse results than Random Importance. This means that ELA importance rank is sensitive to a slight change in the value of the feature. This result is due to the dependency on having appropriate neighborhood points in the data set, and this impact should be reduced when using larger data sets. Finally, the Integrated Gradient obtained worse results than Random Importance under the Stability measure (Fig.~\ref{fig:stability_heatmap}) for the ensemble methods and kNN. We notice from these results that the SR methods obtained very close results to the reference ground-truth (Feynman), meaning that they are as robust as the true expression.

\begin{figure}[t!]
    \begin{subfigure}[b]{\textwidth}
        \centering
        \includegraphics[width=\linewidth]{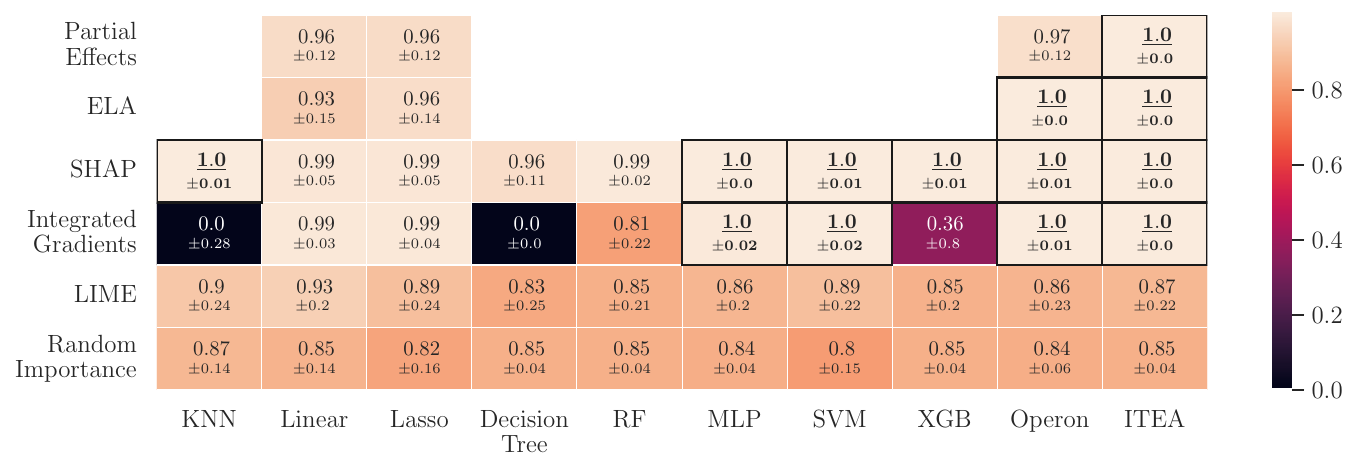}
        \caption{Median Cosine Similarity (greater is better) \textit{heatmap} between local explanations obtained over the regressor and the original Feynman equation.}
        \label{fig:cosine_local_heatmap}
    \end{subfigure} \\%
    \begin{subfigure}[b]{\textwidth}
        \centering
        \includegraphics[width=\linewidth]{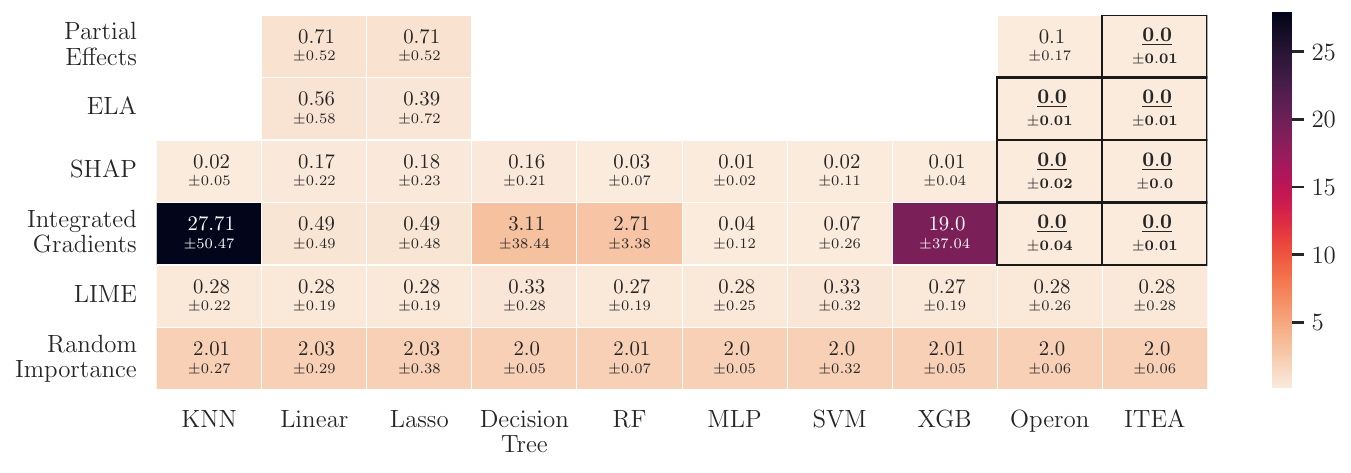}
        \caption{Median NMSE (smaller is better) \textit{heatmap} between local explanations obtained over the regressor and the original Feynman equation.}
        \label{fig:NMSE_local_heatmap}
    \end{subfigure} \\%
    \caption{Local explanation quality metrics \textit{heatmaps} showing the median and standard deviation for each explainer-regression combination. The best values are highlighted with a black border around it, and the color maps always attribute lighter colors to better results.}
    \label{fig:local_quality_heatmaps}
\end{figure}

Fig.~\ref{fig:local_quality_heatmaps} shows the heatmap plots for the Cosine similarity and the NMSE of the local explanations calculated using the ground-truth as a reference point. Considering the cosine similarity, which measures whether the sign of the feature importance is the same as the ground-truth, we can see that almost every combination obtained a value closer to $1.0$, meaning that even an inaccurate model can at least indicate correctly whether a feature has a positive or negative influence to the target feature. The SR methods obtained the best possible results with every explanation method except LIME. When also considering the magnitude of the feature importance, we can see in Fig.~\ref{fig:NMSE_local_heatmap} the SHAP explainer is capable of returning explanations with smaller errors for black-box models. On the other hand, LIME presented a very similar performance for every regressor. Viewing from the regressor side, the SR methods returned the best explanations for every method, except LIME. These results indicate that not only an accurate model is essential for a correct explanation but also that is must be a flexible model capable of capturing the different properties of the generating function.

\subsubsection{Global Explanations}

\begin{figure}[t!]
    \begin{subfigure}[b]{\textwidth}
        \centering
        \includegraphics[width=\linewidth]{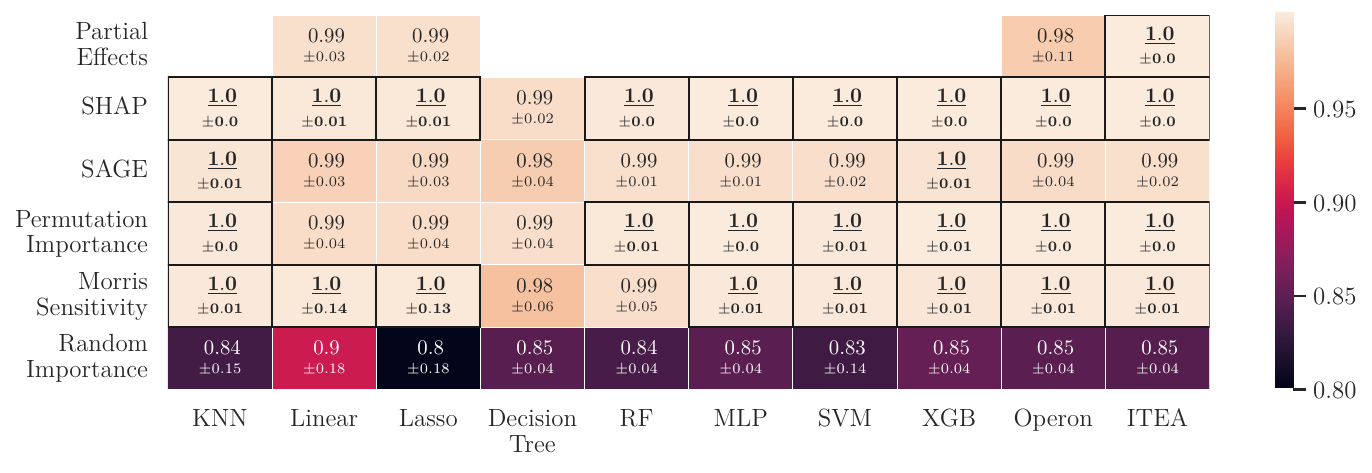}
        \caption{Median Cosine Similarity (greater is better) \textit{heatmap} between global explanations obtained over the regressor and  the original Feynman equation.}
        \label{fig:cosine_global_heatmap}
    \end{subfigure} \\%
    \begin{subfigure}[b]{\textwidth}
        \centering
        \includegraphics[width=\linewidth]{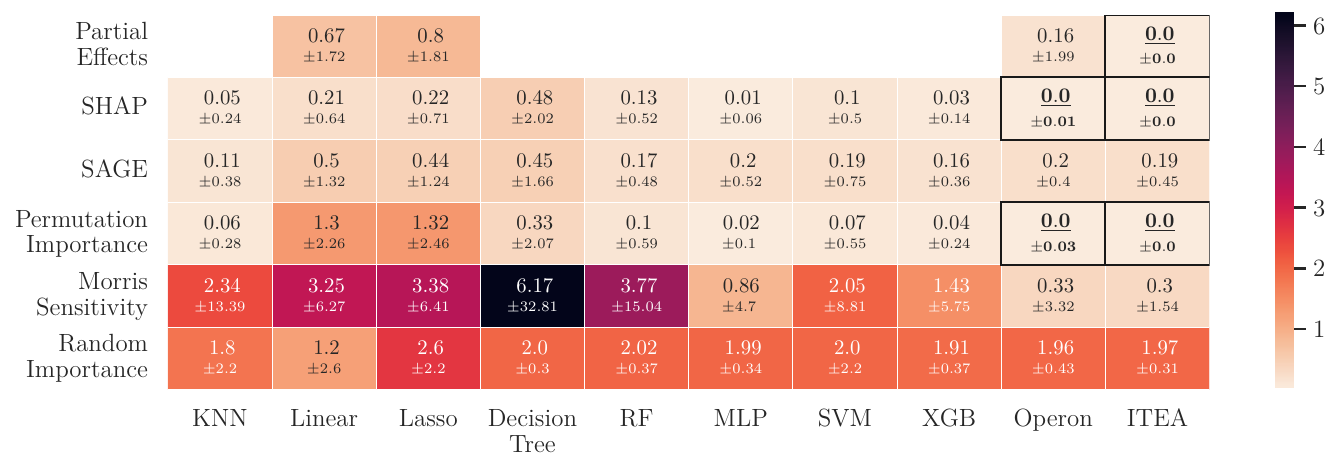}
        \caption{NMSE (smaller is better) \textit{heatmap} between global explanations obtained over the regressor and  the original Feynman equation.}
        \label{fig:NMSE_heatmap}
    \end{subfigure} \\%
    \caption{Global explanation quality metrics \textit{heatmaps} showing the median and standard deviation for each explainer-regression combination. The best values are highlighted with a black border around it, and the color maps always attribute lighter colors to better results.}
    \label{fig:global_quality_heatmaps}
\end{figure}

Fig.~\ref{fig:global_quality_heatmaps} shows the heatmaps for the global explanations. The cosine similarity for the global models is very similar to the local explanations, and most combinations of regressors and explanation models can indicate the correct sign of the effect of each feature. On the other hand, we observed a higher increase in NMSE when comparing SR models against simpler models like Linear and Lasso. The SR methods outperformed other methods with every explainer. Regarding the explainers, Morris Sensitivity is more sensitive to model accuracy, and every other explanation model maintained a small error except for the linear models.

\subsection{Selected benchmark}~\label{sec:selectedresults}

This subsection reports the obtained results with the selected data sets focused on the SR models. These data sets are known to be challenging to any regression model, and, as such, we expect to observe a disagreement in the explanation of the SR models compared to the ground-truth.

\subsubsection{Model accuracy}~\label{sec:modelacc2}

Table~\ref{tab:regression_metrics_comparison} shows the accuracy of ITEA, Operon and MLP models for the selected data sets. We chose MLP as a black-box representative since it obtained the next best results after SR models. We can see from this table that the SR models are still the most accurate among the tested models, but, except for Pagie-1, all algorithms created inaccurate models, as can be seen from the high values for MAE and NMSE. The former metric is more robust to outliers --- as it uses the absolute values---, and the latter tends to alleviate the prediction errors smaller than $1.0$, while punishing prediction errors higher than this value --- since it squares the errors. The SR methods were outperformed by the best black-box model only using the MAE metric on the Korns-12 data set.

In practice, those models would be discarded as they do not capture the relationship of the observation features with the target value. Notice that this creates a situation where we have an unreliable model but will still extract the feature importance information.

\begin{table}[t!]
    \centering
    \caption{Median of MAE and NMSE for the two symbolic regression method and the best performing black-box method for the selected data sets from the literature. The best value for each data within each metric set is highlighted in bold.}
    \label{tab:regression_metrics_comparison}
    \begin{tabular}{@{\extracolsep{5pt}}rc@{\extracolsep{5pt}}c@{\extracolsep{5pt}}cc@{\extracolsep{5pt}}c@{\extracolsep{5pt}}cc@{\extracolsep{5pt}}c@{\extracolsep{5pt}}c@{}}
        \toprule
                   & \multicolumn{3}{c}{MAE}                             & \multicolumn{3}{c}{NMSE} \\
                     \cmidrule{2-4}                                          \cmidrule{5-7}
                   & ITEA              & Operon          & MLP             & ITEA              & Operon           & MLP             \\
                     \cmidrule{1-1} \cmidrule{2-4} \cmidrule{5-7}
                     
        Korns-11   & $7.67$ & $\mathbf{7.26}$ & $7.65$       & $ \mathbf{15.43}$ & $ 32.76$        & $262.80$         \\
        Korns-12   & $0.92$ & $0.88$ & $\mathbf{0.87}$       & $ \mathbf{17.43}$ & $ 30.03$        & $ 30.07$         \\
        Vladislav. & $1.41$ & $\mathbf{0.19}$ & $0.31$       & $\mathbf{1.18}$   & $2.97$          & $1.19$           \\
        Pagie-1    & $\mathbf{0.00}$ & $0.01$ & $0.06$       & $\mathbf{0.00}$   & $\mathbf{0.00}$ & $0.01$           \\
        \bottomrule
    \end{tabular}
\end{table}

The difference in results between ITEA and Operon can be explained by the expression size as reported in Table~\ref{tab:expression_size_and_hit_rate}. We can see that ITEA generated expressions $3$ to $4$ times larger than Operon. Also, as expected, none of the SR algorithms could find a perfect hit in any run.

\begin{table}[t!]
    \centering
    \caption{Median expression size and hit rate (expressed as a percentage since some regressors are non-deterministic) for methods that return mathematical expressions.}
    \label{tab:expression_size_and_hit_rate}
    \begin{tabular}{@{\extracolsep{5pt}}rc@{\extracolsep{5pt}}c@{\extracolsep{5pt}}c@{\extracolsep{5pt}}cc@{\extracolsep{5pt}}c@{}}
    \toprule
               & \multicolumn{4}{c}{Median expression size} & \multicolumn{2}{c}{\# hits} \\
                 \cmidrule{2-5}                               \cmidrule{6-7}
               & Linear & Lasso & ITEA  & Operon & ITEA  & Operon \\ \cmidrule{1-1} \cmidrule{2-5} \cmidrule{6-7}
    Korns-11   & 21.0   & 1.0   & 181.0 & 42.0   & 0\%    & 0\%     \\
    Korns-12   & 16.0   & 1.0   & 180.0 & 56.5   & 0\%    & 0\%     \\
    Vladislav. & 17.0   & 1.0   & 110.5 & 72.0   & 0\%    & 0\%     \\
    Pagie-1    & 1.0    & 1.0   & 92.0  & 24.0   & 0\%    & 0\%     \\ \bottomrule
    \end{tabular}
\end{table}

\subsubsection{Local Explanations}

\def\*#1{\mathbf{#1}}

Table~\ref{tab:local_explanatory_quality} shows a detailed analysis of the performance of each local explainer with ITEA and Operon, focusing on the best performing explanatory methods observed in the heatmap in Fig.~\ref{fig:local_quality_heatmaps}: Integrated Gradients (IG), Partial Effects (PE) and SHAP. We report the two explanation quality measures cosine similarity and the MSE normalized by the variance of the expected explanation (NMSE). Results marked with an $^*$ were not normalized by the original feature importance explanation since every feature has the same importance making the variance zero. 

\begin{table}[t!]
    \centering
    \caption{Median of the quality measures for the selected \textbf{local} explanatory methods with ITEA and Operon.}
    \label{tab:local_explanatory_quality}
    \begin{tabular}{@{\extracolsep{5pt}}rc@{\extracolsep{5pt}}c@{\extracolsep{5pt}}cc@{\extracolsep{5pt}}c@{\extracolsep{5pt}}c@{}}
        \toprule
         \multirow{3}{*}{\large ITEA} &
         \multicolumn{3}{c}{Cosine similarity} & \multicolumn{3}{c}{NMSE} \\
         \cmidrule{2-4}                          \cmidrule{5-7}  
         
                  & IG   & PE   & SHAP & IG   & PE   & SHAP      \\
                    \cmidrule{1-1} \cmidrule{2-4} \cmidrule{5-7} 
                    
        Korns-11  & $\*{1.00}$ & $-0.07$    & $-0.48$ & $50.28$   & $9.00e+14$   & $\*{1.64}$   \\ 
        Korns-12  & $\*{1.00}$ & $0.00$     & $-0.34$ & $1.34e+7$ & $5.61e+17$   & $\*{3.87}$   \\ 
        Vladslav. & $\*{1.00}$ & $0.87$     & $0.91$  & $0.51^*$  & $0.45^*$     & $\*{0.14}^*$ \\
        Pagie-1   & $\*{1.00}$ & $\*{1.00}$ & $-0.61$ & $0.47^*$  & $\*{0.00}^*$ & $0.24^*$     \\ \midrule
        
        \multirow{3}{*}{\large Operon}  &
        \multicolumn{3}{c}{Cosine similarity} & \multicolumn{3}{c}{NMSE} \\
        \cmidrule{2-4}                          \cmidrule{5-7}
        
                   & IG   & PE   & SHAP & IG   & PE   & SHAP      \\ 
                    \cmidrule{1-1} \cmidrule{2-4} \cmidrule{5-7} 
                    
        Korns-11   & $0.25$ & $\*{0.05}$     & $0.07$     & $\*{1.25}$ & $\*{1.25}$   & $1.33$   \\
        Korns-12   & $\*{1.00}$ & $\*{0.00}$ & $-0.07$    & $126.71$   & $3.20$   & $\*{2.99}$   \\
        Vladislav. & $\*{1.00}$ & $0.270$    & $0.93$     & $0.03^*$   & $\*{0.00}^*$ & $0.01^*$ \\ 
        Pagie-1    & $\*{1.00}$ & $0.85$     & $\*{1.00}$ & $0.98^*$   & $\*{0.00}^*$ & $0.06^*$ \\\bottomrule
    \end{tabular}
\end{table}

With the bad prediction accuracy for these problems, we notice a degradation in the feature importance values generated by the explanation models. Both ITEA and Operon found inconsistent directions for their feature importance when using PE and SHAP on Korns-11 and Korns-12 data sets regarding the cosine similarity. Using IG, they found that the sign of importances were partially correlated with the ground-truth for Korns-12. With the Vladislavleva-4 data set, the directions obtained by IG and SHAP were more consistent with the ground-truth, even though this model was also inaccurate. Pagie-1 did not get a perfect score even though they returned perfectly accurate models. This means that, even though the regression models are perfect w.r.t. the validation set, they may not correspond to the exact behavior of the true model.

Partial Effects had a better performance with Operon rather than with ITEA. To investigate this problem, we selected the expressions from ITEA and Operon that had an explanation error close to the median reported in Table~\ref{tab:local_explanatory_quality} for the Korns-11 data set. The selected ITEA model is a sum of cosines given by:

\begin{equation*} \label{eq:itexpr_korns11}
\begin{split}
    \textup{ITExpr}(x, y, z, v, w) = & -0.308\cdot\textup{cos} \left ( \frac{x w^{4}}{y z^{3} v^{4}} \right ) + 0.133\cdot\textup{cos}\left (\frac{y^2 v^3}{x z^3} \right ) \\
     & -0.089\cdot\textup{cos}\left (\frac{z^4 v^3 w^4}{x^2 y^2}\right ) + 0.137\cdot\textup{cos}\left (\frac{y z^2 w^4}{v} \right ) \\
     & -0.121\cdot\textup{cos}\left (\frac{v^4 w^2}{x^4 z}\right ) + 0.175\cdot\textup{cos}\left (\frac{z w^4}{x y^3 v^4}\right ) \\
     & + 0.104 \cdot \textup{cos}\left (\frac{x^3 y z w}{v^4}\right ) -0.098\cdot\textup{cos}\left (\frac{z^4 v^4}{x^2 y^3}\right ) \\
     & + 0.091\cdot \textup{cos}\left (\frac{x z^3 w^4}{y v^4}\right ) + 2.118.
\end{split}
\end{equation*}

The true importance value for this data set for any given point is $[C, 0, 0, 0, 0]$ with $C$ representing the importance value of $x$ at any given point, and this value is usually with a magnitude of $10^4$. 
We can notice that the ITEA expression uses every feature of the data set, not only the single meaningful variable $x$. This means that it can attribute non-zero importance values to noisy variables. At some points, we observed that the partial effect assigned zero importance to $x$ and a large value for one or two other variables, explaining this large error observed in our results. On the other hand, Operon was more competent in selecting the important features for its models, a selected example for this data set is:

\begin{multline*}\label{eq:operon_korns11}
    \textup{OPExpr}(x, y, z, v, w) = 82.931 \\
    -81.160 \cdot \textup{sin}(\textup{tanh}(-3.433 x)^{2^{2^{2^{2^2}}}}  + \textup{cos}(\textup{tanh}(\textup{tanh}(\textup{tanh}(-1.182 z)))^2)),
\end{multline*}

\noindent which significantly reduces the average error as there is only one noisy variable in this expression.

\section{Discussion}~\label{sec:discussion}

The obtained results bring forward some aspects worth considering whenever we seek to extract additional information from the prediction models. These results show additional evidence favoring Symbolic Regression as a gray-box model that can either be interpreted without additional support or using external models that extract the information of interest.

\subsection{Symbolic Regression can have a good trade-off between accuracy and interpretability}

As we reported in Section~\ref{sec:modelacc}, the Symbolic Regression models obtained more accurate models than the black-box models for the Feynman benchmark. Also, about $40\%$ of the ground-truth models were successfully retrieved by at least one of the SR methods. This indicates the possibility of using the symbolic models explicitly when understanding the model behavior. Also, even for the models that do not correspond to the ground-truth, the symbolic models still allowed the extraction of near-perfect feature importance using different explanatory models, as reported in Section~\ref{sec:localexpls}. Additionally, their explanations were as robust as those returned by MLP and SVM. Unlike these models, Random Forest and Gradient Boosting had lower quality explanations and were less robust due to their internal discretization of the feature space. 
Besides these advantages, the symbolic model also allows us, under appropriate conditions, to generate the symbolic partial derivatives of the model. This, in turn, makes it possible to calculate another feature importance measure: the partial effect. With the partial derivative expressions, we can also reduce the estimation error of the Integrated Gradient.

We also found some counterpoints to the above conclusion in Section~\ref{sec:selectedresults} where we tested a set of challenging benchmarks. None of the regression models could generate a reasonable predictor for three out of the four data sets. As an implication of these models, we observed a higher error in the feature importance estimation. The main problem was that all of the predictors could not select the correct features for the model, thus attributing non-zero importance to noisy features. This leads to our next point of discussion.

When comparing the SR methods with the linear methods (Linear and Lasso) when using the regression-specific explainer Partial Effects, it is crucial to notice that linear methods will always hold equal or more stable explanations since the models do not perform any transformation or interaction over the variables; thus small changes will always have a direct and linear effect in the output. The SR methods, however, could approximate the behavior of linear models for explanation robustness, as we can see from the similar reported results for the robustness metrics in Fig.~\ref{fig:local_robustness_heatmaps}.

\subsection{The explanation quality correlates with the model quality}

Following the previous observation, we found that most explanatory models could still return a reasonable explanation even when the prediction models were not perfect (consider, for example, the accuracy of SVM explanations). Nevertheless, a practitioner should ensure that the prediction model has a reasonable quality and be aware of the possible inaccuracies when extracting an explanation. At the bare minimum, the practitioner should follow the standard Machine Learning pipeline of data cleaning, feature engineering, and hyper-parameters optimization for high-stakes prediction tasks before trusting the explanation extracted from the models.

We noticed that we did not perform an extensive grid search in the challenging benchmarks due to computational and time constraints. Having said that, a more careful experiment could improve the SR models quality for these challenging problems. Even when the model is accurate, we still have to be careful about the explanations because some regression models do not comply with the \textit{desiderata} of explanatory methods.

\subsection{Trusting an explanation requires evaluating its robustness and quality w.r.t. the prediction model}
 
Even when the prediction model was accurate, we observed some instability in the explanations as depicted in Fig.~\ref{fig:local_robustness_heatmaps} of Section~\ref{sec:localexpls}. We noticed that the explanations produced by Integrated Gradient, when coupled with decision tree-based models, were unstable since these models generate a step function to approximate non-linear relationships. The explanation at specific points may suffer an enormous change when evaluated at neighborhood points. The practitioner should be aware of these limitations when choosing the right prediction and explanatory methods.

\section{Conclusions}~\label{sec:conclusion}

This paper proposed a benchmark of explanatory methods for regression models using the Feynman data set as a proxy to evaluate these models using a ground-truth as a reference. The main objective was to compare the performance of Symbolic Regression models when coupled with different explainers in contrast with other linear and non-linear regression models coupled with the same explainers.

For this purpose, we have implemented the \textit{iirsBenchmark} as an open-source framework that contains wrapper methods to different regression models, including two symbolic regression and many global and local explanatory methods that return the importance of each feature for the prediction. This framework evaluates each pair regressor-explainer following the robustness and quality criteria. The robustness criteria tests how sensitive an explainer is with slight variations in the data. The \textit{desideratum} is that an explainer will not abruptly change the explanation with small perturbations. The quality of the explanations is evaluated as to whether the direction of the feature importance agrees with the ground-truth and how close they are to the true explanation. 

We have done extensive experiments to evaluate how well Symbolic Regression models fare in this setting. We have found that, specifically for the Feynman data sets, SR was capable of returning accurate models that made it possible to extract the correct explanations even when the returned expression did not correspond to the ground-truth. Compared to the other regressors, SR performance was among the most stable explanations and closer to the ground-truth.

In conclusion, we have found evidence that SR models can return accurate models that correspond to the expected properties and behavior of the true model as captured by the explanatory methods. 

For future work, we intend to expand the coverage for more symbolic regression methods, making it possible to compare different representations for symbolic regression, different coefficient optimization methods, and different evolutionary heuristics. One shortcoming of our experimental methodology is that we only tested problems of the same domain (i.e., physics). We will also add more regression problems from different domains in future work.

We also plan to investigate the role of dimensionality in feature importance explanations. We expect that dimensionality will have a compromise with explanation quality and robustness since larger dimensionalities add more degree of freedom when creating feature importance explanations.

\backmatter

\bmhead{Acknowledgments}

This work was funded by Federal University of ABC (UFABC), Coordenação de Aperfeiçoamento de Pessoal de Nível Superior (CAPES) and Fundação de Amparo à Pesquisa do Estado de São Paulo (FAPESP), grant number 2018/14173-8.

\section*{Declarations}

The authors have no conflicts of interest to declare that are relevant to the content of this article.

\bibliography{bibliography}

\end{document}